%% file: neurips_2025.tex
\documentclass{article}
\usepackage{microtype}
\usepackage{graphicx}
\usepackage{subfigure}
\usepackage{multirow}
\usepackage{booktabs} % for professional tables
\usepackage{stackengine,scalerel}
\usepackage{tcolorbox}
\usepackage{xurl}
\usepackage{enumitem}
\usepackage{hyperref}
\usepackage{wrapfig}

\definecolor{link color}{HTML}{626572}
\hypersetup{
    colorlinks = true,
    linkcolor  = link color,
    filecolor  = link color,
    urlcolor   = link color,
    citecolor  = link color,
}

\usepackage{amssymb}

\DeclareRobustCommand*{\circledbullet}{%
    \mathbin{%
        \ooalign{$\circledcirc$\cr\hidewidth$\bullet$\hidewidth}%
    }%
}

\usepackage{natbib}
\usepackage[final]{neurips_2025}

\usepackage[utf8]{inputenc} % allow utf-8 input
\usepackage[T1]{fontenc}    % use 8-bit T1 fonts
\usepackage{hyperref}       % hyperlinks
\usepackage{url}            % simple URL typesetting
\usepackage{booktabs}       % professional-quality tables
\usepackage{amsfonts}       % blackboard math symbols
\usepackage{nicefrac}       % compact symbols for 1/2, etc.
\usepackage{microtype}      % microtypography
\usepackage{xcolor}         % colors

\usepackage{amsmath}
\usepackage{amssymb}
\usepackage{mathtools}
\usepackage{amsthm}

\usepackage[capitalize,noabbrev]{cleveref}

\theoremstyle{plain}

\theoremstyle{definition}

\theoremstyle{remark}

\usepackage[textsize=tiny]{todonotes}

\input{math_commands}

\title{1000 Layer Networks for Self-Supervised RL: Scaling Depth Can Enable New Goal-Reaching Capabilities}

\author{%
  Kevin Wang \\
  Princeton University \\
  \texttt{kw6487@princeton.edu} \\
  \And
  Ishaan Javali \\
  Princeton University \\
  \texttt{ijavali@princeton.edu} \\
  \And
  Michał Bortkiewicz \\
  Warsaw University of Technology \\
  \texttt{michal.bortkiewicz.dokt@pw.edu.pl} \\
  \And
  Tomasz Trzciński \\
  Warsaw University of Technology, \\
  Tooploox, IDEAS Research Institute \\
  \And
  Benjamin Eysenbach \\
  Princeton University \\
  \texttt{eysenbach@princeton.edu} \\
}

\begin{document}

\maketitle

\begin{abstract}
Scaling up self-supervised learning has driven breakthroughs in language and vision, yet comparable progress has remained elusive in reinforcement learning (RL). In this paper, we study building blocks for self-supervised RL that unlock substantial improvements in scalability, with network depth serving as a critical factor. Whereas most RL papers in recent years have relied on shallow architectures (around 2 -- 5 layers), we demonstrate that increasing the depth up to 1024 layers can significantly boost performance.
Our experiments are conducted in an unsupervised goal-conditioned setting, where no demonstrations or rewards are provided, so an agent must explore (from scratch) and learn how to maximize the likelihood of reaching commanded goals.
Evaluated on simulated locomotion and manipulation tasks, our approach increases performance on the self-supervised contrastive RL algorithm by $2\times$ -- $50\times$, outperforming other goal-conditioned baselines.
Increasing the model depth not only increases success rates but also qualitatively changes the behaviors learned. The project webpage and code can be found here: \href{https://wang-kevin3290.github.io/scaling-crl/}{https://wang-kevin3290.github.io/scaling-crl/}.

% \footnote{Project website: \url{https://wang-kevin3290.github.io/scaling-crl/}} 
%, with more sophisticated behaviors emerging as model capacity grows.
% We additionally find that scaled-up networks unlock the ability to learn representations that capture long-horizon state-action relationships. While our method includes no explicit mechanism for ``stitching,'' we find that increasing network depth increases the capacity to recombine pieces of data to solve unseen tasks.
% , even in the challenging unsupervised goal-conditioned setting without auxiliary rewards or demonstrations.
% , as well as improved combinatorial generalization through the ``stitching'' of states, actions, and goals. 

\end{abstract}

\begin{figure*}[t]
    \centering
    \includegraphics[width=\linewidth]{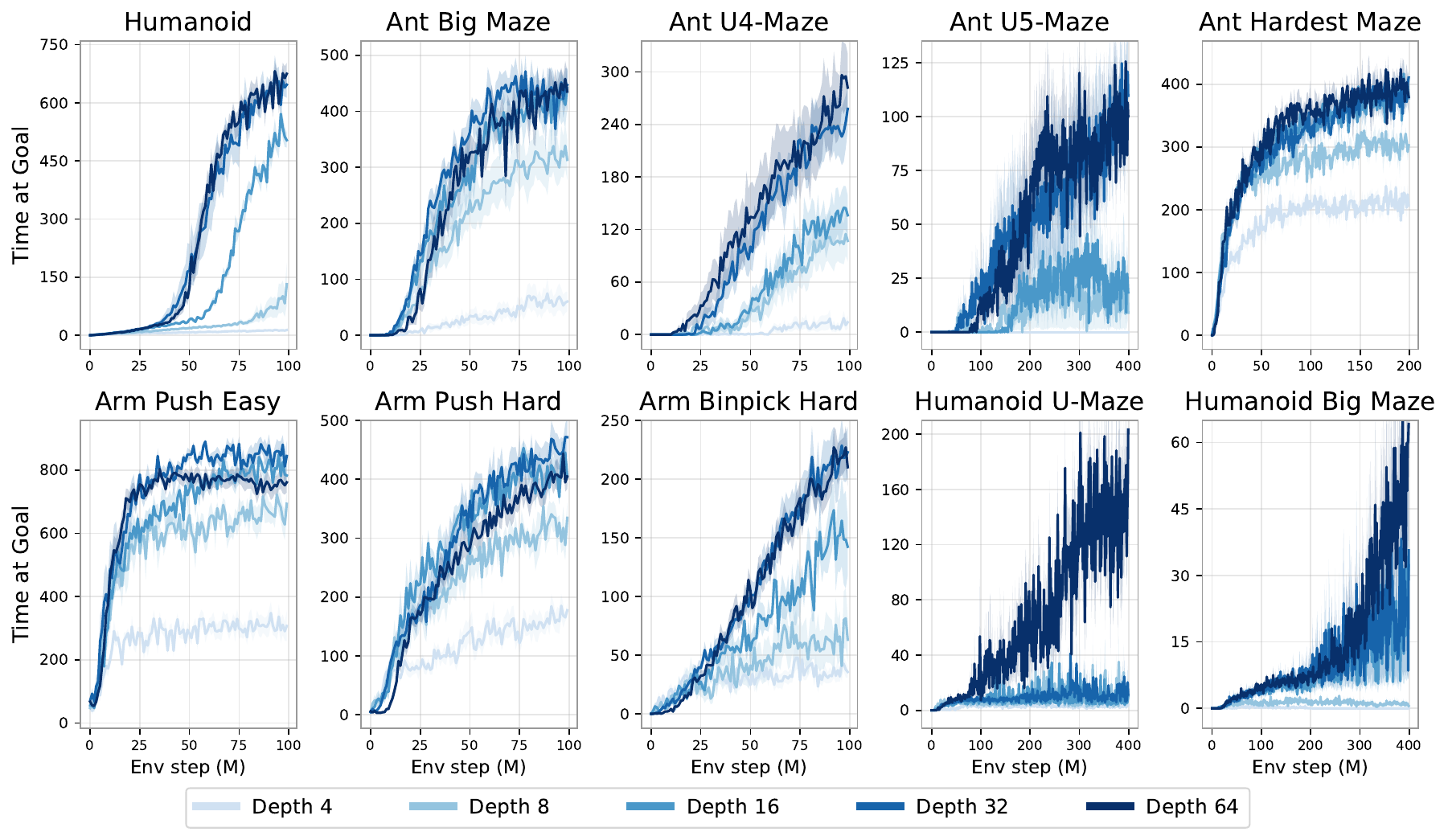}
    \vspace{-1em}
    %https://mydella.princeton.edu/node/della-vis1.princeton.edu/53663/notebooks/scratch/gpfs/kw6487/JaxGCRL/clean_JaxGCRL/wandb_figures.ipynb
    \caption{\footnotesize \textbf{Scaling network depth yields performance gains} across a suite of locomotion, navigation, and manipulation tasks, ranging from doubling performance to 50× improvements on Humanoid-based tasks. Notably, rather than scaling smoothly, performance often jumps at specific critical depths (e.g., 8 layers on Ant Big Maze, 64 on Humanoid U-Maze), which correspond to the emergence of qualitatively distinct policies (see Section \ref{sec:experiments}).}
    \label{fig:main-figure}
    \vspace{-10pt}
\end{figure*}

\section{Introduction}

While scaling model size has been an effective recipe in many areas of machine learning, its role and impact in reinforcement learning (RL) remain unclear. The typical model size for state-based RL tasks is between 2 to 5 layers~\citep{stable-baselines3,huang2022cleanrl}. In contrast, it is not uncommon to use very deep networks in other domain areas; Llama 3~\citep{dubey2024llama} and Stable Diffusion 3~\citep{esser2024scaling} have hundreds of layers.
% \ben{Add rough number of papers at (say) NeurIPS that use small networks; or some other example (e.g., average network size for Atari in Stable Baselines vs for ImageNet in standard library}. 
In fields such as vision~\citep{radford2021learning,zhai2021scaling,dehghani2023scaling} and language~\citep{srivastava2023beyond}, models often only acquire the ability to solve certain tasks once they are larger than a critical scale.
In the RL setting, many researchers have searched for similar emergent phenomena~\citep{srivastava2023beyond}, but these papers typically report only small marginal benefits and typically only on tasks where small models already achieve some degree of success~\citep{naumanBiggerRegularizedOptimistic2024,leeSimBaSimplicityBias2024,farebrotherStopRegressingTraining2024}.
% \ben{I really don't like the word ``breakthrough'' -- let's find a better term that seems less self-aggrandizing} \kevin{Could we not just say "emergent" phenomena here, or is there something else you're looking for?}
A key open question in RL today is whether it is possible to achieve similar jumps in performance by scaling RL networks.

At first glance, it makes sense why training very large RL networks should be difficult: the RL problem provides very few bits of feedback (e.g., only a sparse reward after a long sequence of observations), so the ratio of feedback to parameters is very small. The conventional wisdom~\citep{lecun2016predictive}, reflected in many recent models~\citep{radford2018improving, chen2020big, goyal2019scaling}, has been that large AI systems must be trained primarily in a self-supervised fashion and that RL should only be used to finetune these models. Indeed, many of the recent breakthroughs in other fields have been primarily achieved with \textit{self-supervised} methods, whether in computer vision~\citep{caron2021emerging,radford2021learning,liu2024visual}, NLP~\citep{srivastava2023beyond}, or multimodal learning~\citep{zong2024selfsupervisedmultimodallearningsurvey}. Thus, if we hope to scale reinforcement learning methods, self-supervision will likely be a key ingredient.

In this paper, we will study \emph{building blocks} for scaling reinforcement learning. Our first step is to rethink the conventional wisdom above: ``reinforcement learning'' and ``self-supervised learning'' are not diametric learning rules, but rather can be married together into self-supervised RL systems that explore and learn policies without reference to a reward function or demonstrations~\citep{eysenbach2021clearning,eysenbach2022contrastive,leeMultiGameDecisionTransformers2022}.
In this work, we use one of the simplest self-supervised RL algorithms, contrastive RL (CRL)~\citep{eysenbach2022contrastive}.
The second step is to recognize the importance of increasing data availability. We will do this by building on recent GPU-accelerated RL frameworks~\citep{makoviychuk2021isaac, rutherford2023jaxmarl, rudin2022learning, bortkiewicz2024accelerating}.
The third step is to increase network depth, using networks that are up to $100\times$ deeper than those typically found in prior work. Stabilizing the training of such networks will require incorporating architectural techniques from prior work, including residual connections \citep{he2015deep}, layer normalization~\citep{ba2016layer}, and Swish activation~\citep{DBLP:conf/iclr/RamachandranZL18}. Our experiments will also study the relative importance of batch size and network width.

The primary contribution of this work is to show that a method that integrates these building blocks into a single RL approach exhibits strong scalability:

\begin{itemize}[itemsep=2pt, topsep=2pt, parsep=0pt, partopsep=0pt]
    \item \textbf{Empirical Scalability:} 
    We observe a significant performance increase, more than $20\times$ in half of the environments and outperforming other standard goal-conditioned baselines. These performance gains correspond to qualitatively distinct policies that emerge as the scale increases.
    \item \textbf{Scaling Depth in Network Architecture:} While many prior RL works have primarily focused on increasing network width, they often report limited or even negative returns when expanding depth~\citep{leeSimBaSimplicityBias2024, naumanBiggerRegularizedOptimistic2024}. In contrast, our approach unlocks the ability to scale along the axis of depth, yielding performance improvements that surpass those from scaling width alone (see Sec.~\ref{sec:experiments}). 
    \item \textbf{Empirical Analysis}:  We conduct an extensive analysis of the key components in our scaling approach, uncovering critical factors and offering new insights.

\end{itemize}
We anticipate that future research may build on this foundation by uncovering additional building blocks.
% We summarize our contributions:
% \begin{itemize}
%     \vspace{-5pt}
%     \setlength\itemsep{0.2em}
%     \item \textbf{Effective scaling of state-based RL} - we show a steady performance increase and new behaviors learned while scaling up actor and critic architectures. 
%     % \michal{Both in offline and online?}
%     \item \textbf{Emergent properties unlocked through increased depth} - we demonstrate the first signs of these properties, which have so far been observed primarily in vision and language models, such as critical batch size increase with the number of model parameters.
%     \item \textbf{Extensive empirical analysis} - we offer several learnings of scaling up CRL, including the relation of performance to actor and critic size, the impact of other hyperparameters, and generalization correlation with model depth. 
% \end{itemize}

\section{Related Work}
% We build upon recent advances in GCRL, self-supervised RL, and architecture scaling in RL, demonstrating substantial performance improvement and emergent behaviors in CRL due to network depth scaling.

Natural Language Processing (NLP) and Computer Vision (CV) have recently converged in adopting similar architectures (i.e. transformers) and shared learning paradigms (i.e self-supervised learning), which together have enabled transformative capabilities of large-scale models~\citep{vaswani2017attention,srivastava2023beyond,zhai2021scaling,dehghani2023scaling,wei2022emergent}. In contrast, achieving similar advancements in reinforcement learning (RL) remains challenging. Several studies have explored the obstacles to scaling large RL models, including parameter underutilization~\citep{obandoceron2024mixtures}, plasticity and capacity loss~\citep{lyle2024disentangling,lyle2022understanding}, data sparsity~\citep{andrychowicz2017hindsight,lecun2016predictive}, and training instabilities~\citep{ota2021training,henderson2018deep,van2018deep,DBLP:conf/icml/NaumanBMTOC24}.
As a result, current efforts to scale RL models are largely restricted to specific problem domains, such as imitation learning~\citep{tuylsScalingLawsImitation2024}, multi-agent games~\citep{neumann2022scaling}, language-guided RL~\citep{driess2023palme,ahn2022say}, and discrete action spaces~\citep{obandoceron2024mixtures,DBLP:conf/icml/SchwarzerOCBAC23}.
% To date, no unified set of best practices for scaling RL models exists.

Recent approaches suggest several promising directions, including new architectural paradigms~\citep{obandoceron2024mixtures}, distributed training approaches~\citep{ota2021training,espeholt2018impala}, distributional RL~\citep{DBLP:conf/iclr/KumarAGTL23}, and distillation~\citep{team2023human}. Compared to these approaches, our method makes a simple extension to an existing self-supervised RL algorithm. The most recent works in this vein include \citet{leeSimBaSimplicityBias2024} and \citet{naumanBiggerRegularizedOptimistic2024}, which leverage residual connections to facilitate the training of wider networks. These efforts primarily focus on network width, noting limited gains from additional depth, thus both works use architectures with only four MLP layers. In our method, we find that scaling width indeed improves performance (\cref{sec:what_matter}); however, our approach also enables scaling along depth, proving to be more powerful than width alone.

One notable effort to train deeper networks is described by \citet{farebrotherStopRegressingTraining2024}, who cast value-based RL into a classification problem by discretizing the TD objective into a categorical cross-entropy loss. This approach draws on the conjecture that classification-based methods can be more robust and stable and thus may exhibit better scaling properties than their regressive counterparts~\citep{torgo1996regression,farebrotherStopRegressingTraining2024}.
The CRL algorithm that we use effectively uses a cross-entropy loss as well~\citep{eysenbach2022contrastive}. Its InfoNCE objective is a generalization of the cross-entropy loss, thereby performing RL tasks by effectively classifying whether current states and actions belong to the same or different trajectory that leads toward a goal state. In this vein, our work serves as a second piece of evidence that classification, much like cross-entropy's role in the scaling success in NLP, could be a potential building block in RL.

\section{Preliminaries}

This section introduces notation and definitions for goal-conditioned RL and contrastive RL. Our focus is on online RL, where a replay buffer stores the most recent trajectories, and the critic is trained in a self-supervised manner.

\paragraph{Goal-Conditioned Reinforcement Learning}
We define a goal-conditioned MDP as tuple $\mathcal{M}_g = (\mathcal{S}, \mathcal{A}, p_0, p, p_g, r_g, \gamma)$, where the agent interacts with the environment to reach arbitrary goals~\citep{kaelbling1993learning,andrychowicz2017hindsight,blier2021learning}. At every time step $t$, the agent observes state $s_t \in \mathcal{S}$ and performs a corresponding action $a_t \in \mathcal{A}$. The agent starts interaction in states sampled from $p_0(s_0)$, and the interaction dynamics are defined by the transition probability distribution $p(s_{t+1} \mid s_{t}, a_{t})$. Goals $g \in \mathcal{G}$ are defined in a goal space $\mathcal{G}$, which is related to $\mathcal{S}$ via a mapping $f: \mathcal{S} \to \mathcal{G}$. For example, $\mathcal{G}$ may correspond to a subset of state dimensions. The prior distribution over goals is defined by $p_g(g)$. The reward function is defined as the probability density of reaching the goal in the next time step $r_g(s_{t}, a_{t}) \triangleq (1-\gamma)p(s_{t+1} = g \mid s_t, a_t)$, with discount factor $\gamma$.

% in GCRL depends on the state, action, and the commanded goal from the distribution $p_g\left(g\right)$. 

In this setting, the goal-conditioned policy $\pi(a \mid s, g)$ receives both the current observation of the environment as well as a goal. We define the discounted state visitation distribution as
$p^{\pi\left(\cdot \mid \cdot, g\right)}_{\gamma}(s) \triangleq (1-\gamma)\sum_{t=0}^{\infty} \gamma^{t} p^{\pi\left(\cdot \mid \cdot, g\right)}_{t}(s)$, where $p_t^\pi(s)$ is the probability that policy $\pi$ visits $s$ after exactly $t$ steps, when conditioned with $g$. This last expression is precisely the $Q$-function of the policy $\pi(\cdot  \mid \cdot , g)$ for the reward $r_{g}$: $Q^{\pi}_{g}(s,a) \triangleq p^{\pi\left(\cdot \mid \cdot, g\right)}_{\gamma}(g \mid s,a)$.
% \michal{Finish} Thus, a goal-conditioned policy $\pi(a \mid s, g)$ receives both the current observation of the environment and a goal $g$. 
The objective is to maximize the expected reward: 
\begin{equation}
    \max _\pi \mathbb{E}_{p_0(s_0), p_g\left(g\right), \pi\left(\cdot \mid \cdot, g\right)}\left[\sum_{t=0}^{\infty} \gamma^t r_g\left(s_t, a_t\right)\right].
\end{equation}
% \ben{Add definition of $r_g(s, a)$. See how we did this in the Accelerating paper.}
% \ben{Add definition of the Q-function here.}
% \ben{Add definition of $p_\gamma$ here.}

\paragraph{Contrastive Reinforcement Learning.}

% {\color{orange}Potential ordering of this section:
% \begin{itemize}
%     \item Our experiments will use contrastive RL algorithm [cite] to solve goal-conditioned problems. Contrastive RL is an actor critic method; we will use $f(s, a, g)$ to denote the critic and $\pi(a \mid s, g)$ to denote the policy. The critic is trained with the InfoNCE objective [cite] EQUATION.
%     The actor is trained to maximize the critic EQUATION.
% \end{itemize}
% }

Our experiments will use the contrastive RL algorithm~\citep{eysenbach2022contrastive} to solve goal-conditioned problems. Contrastive RL is an actor-critic method; we will use $f_{\phi, \psi}(s, a, g)$ to denote the critic and $\pi_{\theta}(a \mid s, g)$ to denote the policy. The critic is parametrized with two neural networks that return state, action pair embedding $\phi(s,a)$ and goal embedding $\psi(g)$. The critic's output is defined as the $l^2$-norm between these embeddings: $f_{\phi, \psi}(s,a,g) = \|\phi(s,a) - \psi(g)\|_2$.
% We frame goal-reaching as an inference problem \citep{borsa2019universal,barreto2022successor,blier2021learning,eysenbach2022contrastive} namely, given the current state $s$ and a desired goal $g$, we aim to determine the most likely action $a$ that will lead us to that goal. In this setup, the critic is defined as a classification problem \citep{eysenbach2022contrastive,eysenbach2021clearning,zheng2023contrastive,zheng2024stabilizing,myers2024learning,bortkiewicz2024accelerating}.
% In summary, the critic is modeled as a state-action-goal Q-function $f(s,a,g)$, which estimates the likelihood of reaching the goal and evaluates how different actions impact this likelihood: 
% \begin{equation}
% 	f(s,a,g) \propto p^{\pi}_{\gamma}(g \mid s,a) = Q^{\pi}_{g}(s,a).
% 	\label{eq:optimality_condition}
% \end{equation}
The critic is trained with the InfoNCE objective~\citep{sohn2016improved} as in previous works~\citep{eysenbach2022contrastive,eysenbach2021clearning,zheng2023contrastive,zheng2024stabilizing,myers2024learning,bortkiewicz2024accelerating}. Training is conducted on batches $\mathcal{B}$, where $\textcolor{positive}{s_i, a_i, g_i}$ represent the state, action, and goal (future state) sampled from the same trajectory, while $\textcolor{negative}{g_j}$ represents a goal sampled from a different, random trajectory. The objective function is defined as:
\begin{equation*}
  \min_{\phi, \psi}\mathbb{E}_{\mathcal{B}}\left[-\sum\nolimits_{\textcolor{positive}{i=1}}^{|\mathcal{B}|} \log \biggl( {\frac{e^{f_{\phi,\psi}(\textcolor{positive}{s_{i}},\textcolor{positive}{a_{i}},\textcolor{positive}{g_{i}})}}{\sum\nolimits_{\textcolor{negative}{j=1}}^{K} e^{f_{\phi,\psi}(\textcolor{positive}{s_{i}},\textcolor{positive}{a_{i}},\textcolor{negative}{g_{j}})}}} \biggr) \right].  
\end{equation*}

The policy $\pi_\theta(a \mid s, g)$ is trained to maximize the critic:
% To extract goal-conditioned policy $\pi_\theta(a|s,g)$, we optimize the critic $f_{\phi, \psi}$ with respect to $\theta$ that parametrizes policy:
\begin{equation*}
	\label{eq:policy_loss}
	\max_{\pi_\theta} \mathbb{E}_{\substack{p_0(s_0), p(s_{t+1} \mid s_{t}, a_{t}),\\ p_g\left(g\right), \pi_\theta(a \mid s,g)}} \left[ f_{\phi,\psi}(s,a,g)\right].
\end{equation*}

\paragraph{Residual Connections}

\begin{wrapfigure}[18]{R}{0.48\linewidth}
    \centering
    % \vspace{-3pt}
    \includegraphics[width=\linewidth]{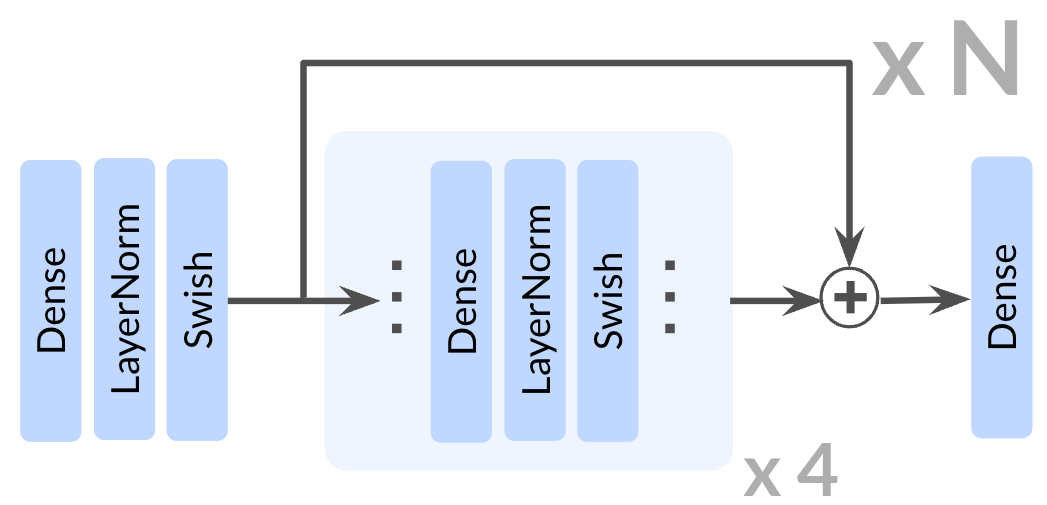}
    \vspace{-0.5em}
    \caption{\footnotesize \textbf{Architecture.} Our approach integrates residual connections into both the actor and critic networks of the Contrastive RL algorithm. The depth of this residual architecture is defined as the total number of Dense layers across the residual blocks, which, with our residual block size of 4, equates to $4N$.}
    \label{fig:architecture}
    \vspace{-20pt}
\end{wrapfigure}
We incorporate residual connections~\citep{he2015deep} into our architecture, following their successful use in RL~\citep{farebrotherStopRegressingTraining2024,leeSimBaSimplicityBias2024,naumanBiggerRegularizedOptimistic2024}. A residual block transforms a given representation $\mathbf{h}_i$ by adding a learned residual function $F_i(\mathbf{h}_i)$ to the original representation. Mathematically, this is expressed as:
% A residual block applied on a representation $h_i$ transforms the representation as:
\begin{equation*}
    \mathbf{h}_{i+1}=\mathbf{h}_i+F_i\left(\mathbf{h}_i\right)
\end{equation*}
where $\mathbf{h}_{i+1}$ is the output representation, $\mathbf{h}_i$ is the input representation, and $F_i(\mathbf{h}_i)$ is a transformation learned through the network (e.g., using one or more layers). The addition ensures that the network learns modifications to the input rather than entirely new transformations, helping to preserve useful features from earlier layers. Residual connections improve gradient propagation by introducing shortcut paths~\citep{he16identity,veit2016residual}, enabling more effective training of deep models. 

\section{Experiments}
\label{sec:experiments}
% In this section, we detail the experiments conducted to evaluate the scalability and effectiveness of our approach across a diverse set of tasks and configurations. First, we describe the experimental setup and environments used in \cref{sec:setup}. Next, in \cref{sec:scaling_main} we present our main results on performance scaling with depth of CRL agent, and follow with demonstrations of emergent behaviors that occur when training CRL with deep architectures in \cref{sec:emergent}. \textit{Second}, in \cref{sec:what_matter} we provide the empirical evidence that scaling happens only for self-supervised RL methods and demonstrate critical components for scaling RL with depth. \textit{Lastly}, in \cref{sec:why_scaling_happens} we investigate the representations of CRL in both online and offline settings to explain why scaling occurs.
% % \footnote{Project website with code: \url{https://wang-kevin3290.github.io/scaling-crl/}} 
% % For additional details and extended ablation studies, please refer to Appendix \ref{sec:appx_exps}.

\subsection{Experimental Setup}
\label{sec:setup}

\paragraph{Environments.} All RL experiments use the JaxGCRL codebase~\citep{bortkiewicz2024accelerating}, which facilitates fast online GCRL experiments based on Brax~\citep{freeman2021brax} and MJX~\citep{todorov2012mujoco} environments. 
% \michal{Should we indicate the cleanRL physics differences?} 
The specific environments used are a range of locomotion, navigation, and robotic manipulation tasks, for details see~\cref{app:technical_details}. We use a sparse reward setting, with $r=1$ only when the agent is in the goal proximity. For evaluation, we measure the number of time steps (out of 1000) that the agent is near the goal.
% All algorithms are instantiated in the goal-conditioned setting with no auxiliary reward, with $r=1$ when the agent is at the goal state and $r=0$ otherwise. Unless otherwise noted, the success metric used in this paper denotes the average time near goal per evaluation episode.
When reporting an algorithm's performance as a single number, we compute the average score over the last five epochs of training.

\paragraph{Architectural Components}
We employ residual connections from the ResNet architecture~\citep{he2015deep}, with each residual block consisting of four repeated units of a Dense layer, a Layer Normalization~\citep{ba2016layer} layer, and Swish activation~\citep{DBLP:conf/iclr/RamachandranZL18}. We apply the residual connections immediately following the final activation of the residual block, as shown in \cref{fig:architecture}. In this paper, we define the depth of the network as the total number of Dense layers across all residual blocks in the architecture. In all experiments, the depth refers to the configuration of the actor network and both critic encoder networks, which are scaled jointly, except for the ablation experiment in \cref{sec:actor_vs_critic}.

% \begin{figure}
%     \centering
%     \includegraphics[width=1\linewidth]{figures/architecture_flat.png}
%     \caption{Enter Caption}
%     \label{fig:enter-label}
% \end{figure}

% \begin{figure}[t]
%     \centering
%     % \includegraphics[width=1\linewidth, angle=90]{figures/architecture_flat.png}
%     \includegraphics[width=0.7\linewidth]{architecture.png}
%     \vspace{-0.5em}
%     \caption{\textbf{Architecture.} Our approach integrates residual connections into both the actor and critic networks of the Contrastive RL (CRL) algorithm. The depth of this residual architecture is defined as the total number of Dense layers across the residual blocks, which with our residual block size of 4, equates to $4N$.}
%     \label{fig:architecture}
% \end{figure}

% \subsection{Depth Scaling in Contrastive RL}
\subsection{Scaling Depth in Contrastive RL}
\label{sec:scaling_main}

% \paragraph{Numerical Gains} Previous literature has explored scaling actor and critic depths up to 4 layers, with little improvement in performance~\citep{leeSimBaSimplicityBias2024,naumanBiggerRegularizedOptimistic2024}.
% Breaking with previous works, we show that extremely deep actor and critic architectures are feasible and effective for state-based RL. 

We start by studying how increasing network depth can increase performance.
Both the JaxGCRL benchmark and relevant prior work~\citep{leeSimBaSimplicityBias2024,naumanBiggerRegularizedOptimistic2024,zheng2024stabilizing} use MLPs with a depth of 4, and as such we adopt it as our baseline. In contrast, we will study networks of depth 8, 16, 32, and 64. The results in \cref{fig:main-figure} demonstrate that deeper networks achieve significant performance improvements across a diverse range of locomotion, navigation, and manipulation tasks. Compared to the 4-layer models typical in prior work, deeper networks achieve $2 - 5 \times$ gains in robotic manipulation tasks, over $20\times$ gains in long-horizon maze tasks such as Ant U4-Maze and Ant U5-Maze, and over $50\times$ gains in humanoid-based tasks. The full table of performance increases up to depth 64 is provided in \cref{tab:performance_metrics}.

In \cref{fig:baselines}, we present results the same 10 environments, but compared against SAC, SAC+HER, TD3+HER, GCBC, and GCSL. Scaling CRL leads to substantial performance improvements, outperforming all other baselines in 8 out of 10 tasks. The only exception is SAC on the Humanoid Maze environments, where it exhibits greater sample efficiency early on; however, scaled CRL eventually reaches comparable performance. These results highlight that scaling the depth of the CRL algorithm enables state-of-the-art performance in goal-conditioned reinforcement learning.% 8 out of 10 tasks.} We compare CRL against five baselines, with SAC+HER and SAC ranking as the second and third best methods, respectively.}

%     \label{fig:baselines}    % \vspace{-1em}
% \end{figure}

% In preliminary experiments, we also evaluated depth scaling in the offline goal-conditioned setting using OGBench ~\citep{park2024ogbench}. We found no evidence that increasing the network depth of CRL improves performance in this offline setting. 
% % Reproducing the baseline CRL results (depth 4, no residual connections), we observed that deeper networks consistently reduced success rates across three environments, both with and without residual connections.
% To further investigate this, we conducted ablations: \textit{(1)} scaling critic depth while holding the actor at 4 or 8 layers, and \textit{(2)} applying cold initialization to the final layers of the critic encoders~\citep{zheng2024stabilizing}. In all cases, baseline depth 4 networks had the highest success. This finding hints that the reason why scaling depth is useful in the online setting might be partially because it aids in exploration.

\subsection{Emergent Policies Through Depth}
\label{sec:emergent}

\begin{wrapfigure}{r}{0.5\textwidth}
    \centering
    \vspace{-10pt}
    \includegraphics[width=\linewidth]{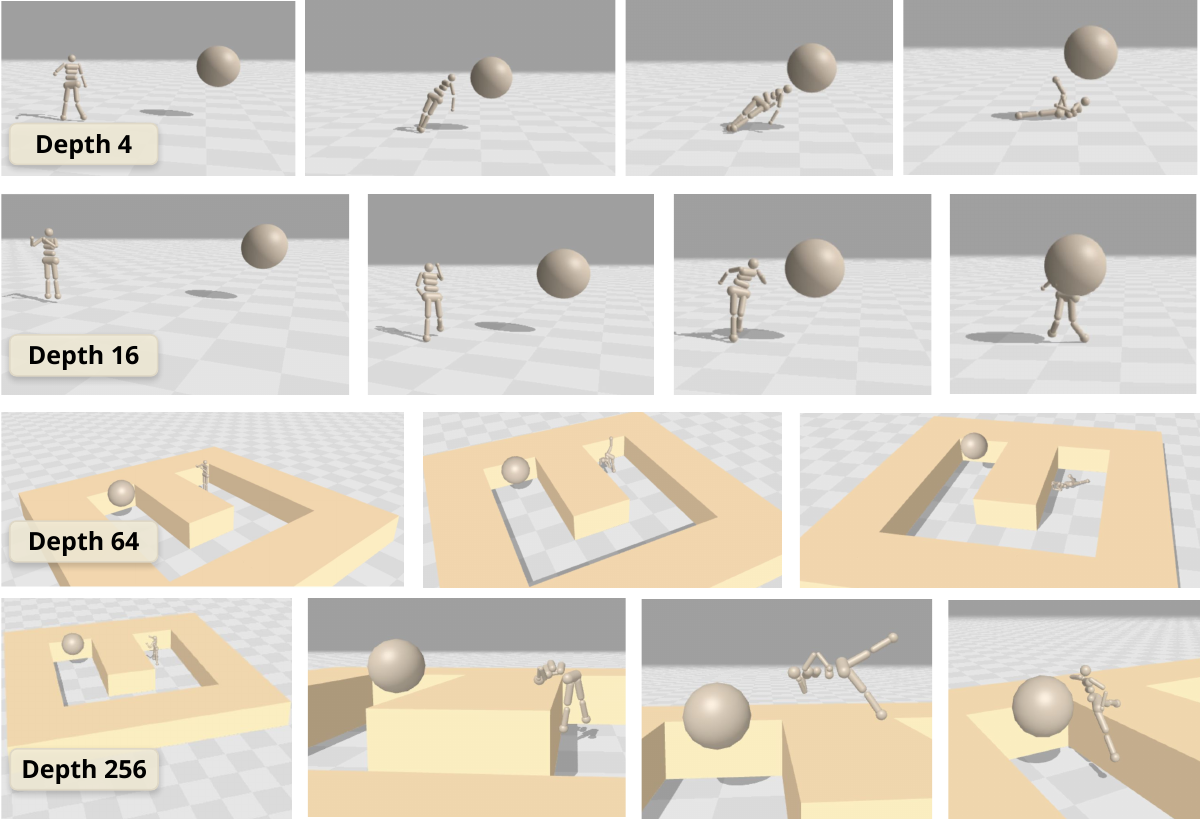}
    \vspace{-10pt}
    \caption{\footnotesize \textbf{Increasing depth results in new capabilities:}
    \textbf{Row 1}: A depth-4 agent collapses and throws itself toward the goal.
    \textbf{Row 2}: A depth-16 agent walks upright.
    \textbf{Row 3}: A depth-64 agent struggles and falls.
    \textbf{Row 4}: A depth-256 agent vaults the wall acrobatically.}
    \label{fig:emergent}
    \vspace{-10pt}
\end{wrapfigure}
A closer examination of the results from the performance curves in \cref{fig:main-figure} reveals a notable pattern: instead of a gradual improvement in performance as depth increases, there are pronounced jumps that occur once a \textit{critical depth} threshold is reached (also shown in \cref{fig:shallow-networks}).
The critical depths vary by environment, ranging from 8 layers (e.g. Ant Big Maze) to 64 layers in the Humanoid U-Maze task, with further jumps occurring even at depths of 1024 layers (see the Testing Limits section, \cref{sec:testing_limits}).
% The critical depth varies by environment, ranging from 8 layers (e.g. Ant Big Maze) to 64 layers in the Humanoid U-Maze task (see the Testing Limits section, \cref{sec:testing_limits}).

% \begin{wrapfigure}{r}{0.5\textwidth}
%     \centering
%     \vspace{-1.5em}
%     \includegraphics[width=0.48\textwidth]{figures/shallowF.png}
%     \vspace{-1em}
%     \caption{\footnotesize \textbf{Critical depth and residual connections.}
%     Incrementally increasing the depth of shallow networks results in marginal performance gains \textit{(left)}. However, when the network depth reaches a critical threshold, performance improves dramatically \textit{(right)}. Residual connections are essential to stabilize the training of large networks in order to fully leverage the benefits of greater depth.}
%     \label{fig:shallow-networks}
% \end{wrapfigure}

Prompted by this observation, we visualized the learned policies at various depths and found qualitatively distinct skills and behaviors exhibited. This is particularly pronounced in the humanoid-based tasks, as illustrated in \cref{fig:emergent}. Networks with a depth of 4 exhibit rudimentary policies where the agent either falls or throws itself toward the target. Only at a critical depth of 16 does the agent develop the ability to walk upright into the goal. 
In the Humanoid U-Maze environment, networks of depth 64 struggle to navigate around the intermediary wall, collapsing on the ground. Remarkably at a depth of 256, the agent learns unique behaviors on Humanoid U-Maze. These behaviors include folding forward into a leveraged position to propel itself over walls and shifting into a seated posture over the intermediary obstacle to worm its way toward the goal (one of these policies is illustrated in the fourth row of \cref{fig:emergent}). To the best of our knowledge, this is the first goal-conditioned approach to document such behaviors on the humanoid environment.

\subsection{What Matters for CRL Scaling}
\label{sec:what_matter}

\paragraph{Width vs. Depth}

\begin{wrapfigure}{r}{0.5\textwidth}
    \centering
    \vspace{-1em}
    \includegraphics[width=0.48\textwidth]{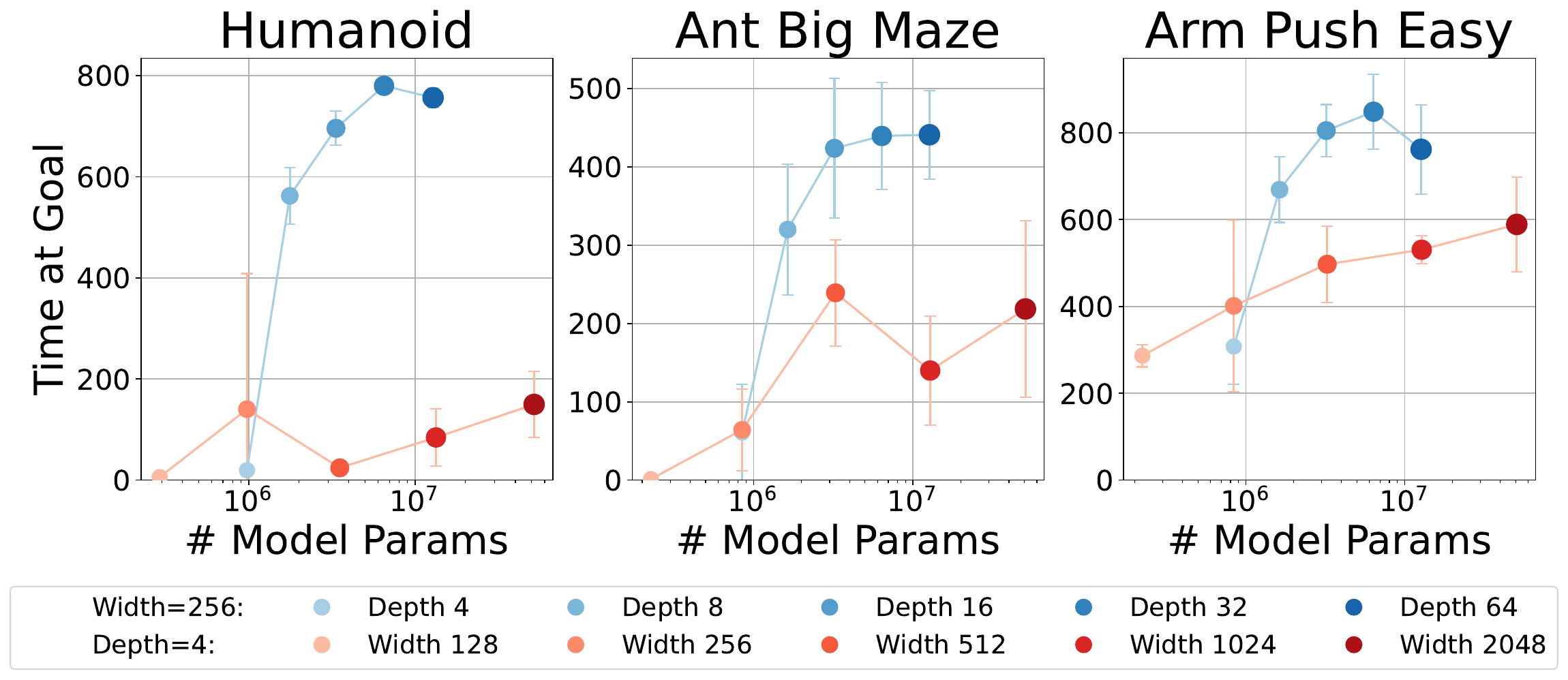}
    \vspace{-5pt}
    \caption{\footnotesize \textbf{Scaling network width vs. depth}. 
    Here, we reflect findings from previous works~\citep{leeSimBaSimplicityBias2024,naumanBiggerRegularizedOptimistic2024} which suggest that increasing network width can enhance performance. However, in contrast to prior work, our method is able to scale depth, yielding more impactful performance gains. For instance, in the Humanoid environment, raising the width to 2048 (depth=4) fails to match the performance achieved by simply doubling the depth to 8 (width=256). The comparative advantage of scaling depth is more pronounced as the observational dimensionality increases.}
    \label{fig:network-width}
    \vspace{-10pt}
\end{wrapfigure}

Past literature has shown that scaling network width can be effective ~\citep{leeSimBaSimplicityBias2024,naumanBiggerRegularizedOptimistic2024}.
In \cref{fig:network-width}, we find that scaling width is also helpful in our experiments: wider networks consistently outperform narrower networks (depth held constant at 4). However, depth seems to be a more effective axis for scaling: simply doubling the depth to 8 (width held constant at 256) outperforms the widest networks in all three environments. The advantage of depth scaling is most pronounced in the Humanoid environment (observation dimension 268), followed by Ant Big Maze (dimension 29) and Arm Push Easy (dimension 17), suggesting that the comparative benefit may increase with higher observation dimensionality.
%Depth 4, 4096 width: 1101824+50343936+262208=51605968
%Depth 16, 256 width: 69120+986880+16448=1072448

Note additionally that the parameter count scales linearly with width but quadratically with depth. For comparison, a network with 4 MLP layers and 2048 hidden units has roughly 35M parameters, while one with a depth of 32 and 256 hidden units has only around 2M. Therefore, when operating under a fixed FLOP compute budget or specific memory constraints, depth scaling may be a more computationally efficient approach to improving network performance.
% a fixed compute budget of FLOPs or specific memory constraints, depth scaling may offer even greater benefits.

\paragraph{Scaling the Actor vs. Critic Networks} 
\label{sec:actor_vs_critic}

\begin{figure}[t]
    \centering
    \begin{minipage}[t]{0.48\textwidth}
        \centering
        \includegraphics[width=\linewidth]{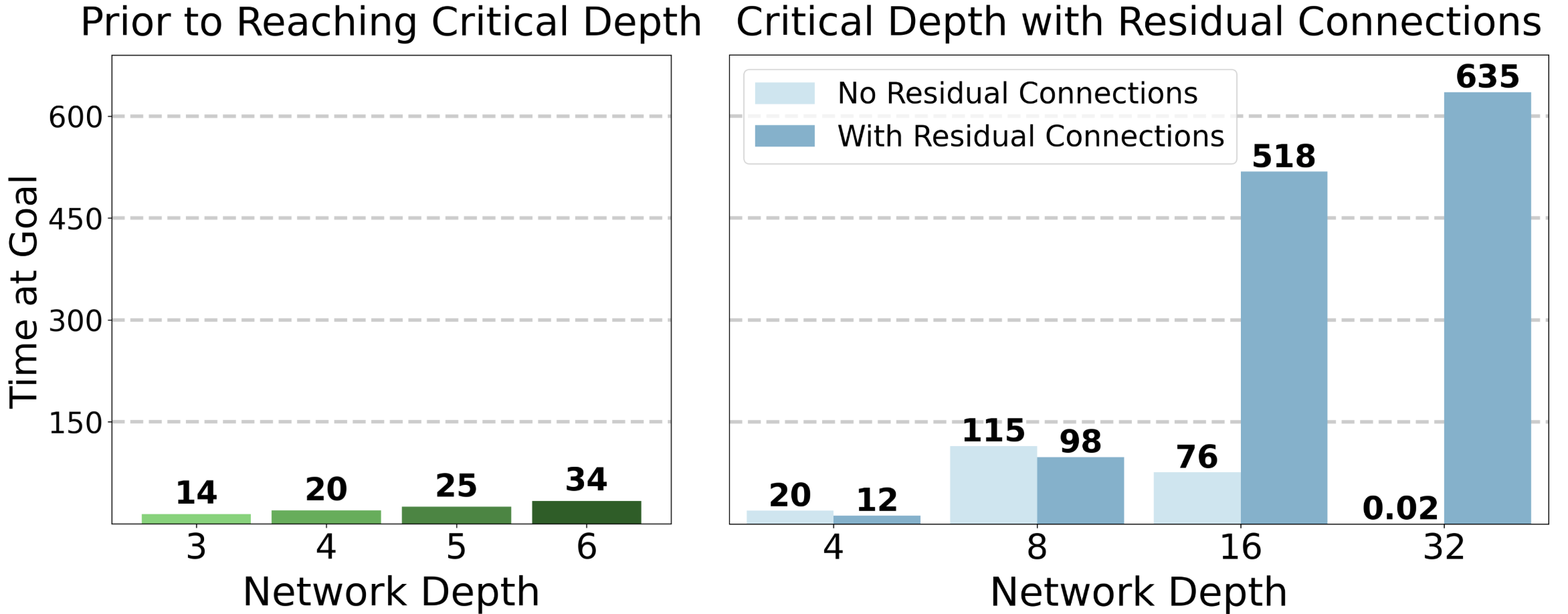}
        \vspace{-0.5em}
        \caption{\footnotesize \textbf{Critical depth and residual connections.}
        Incrementally increasing depth results in marginal performance gains \textit{(left)}. However, once a critical threshold is reached, performance improves dramatically \textit{(right)} for networks with residual connections.}
        % .  are essential to stabilize training and fully leverage deep architectures.}
        \label{fig:shallow-networks}
    \end{minipage}
    \hfill
    \begin{minipage}[t]{0.48\textwidth}
        \centering
        \includegraphics[width=\linewidth]{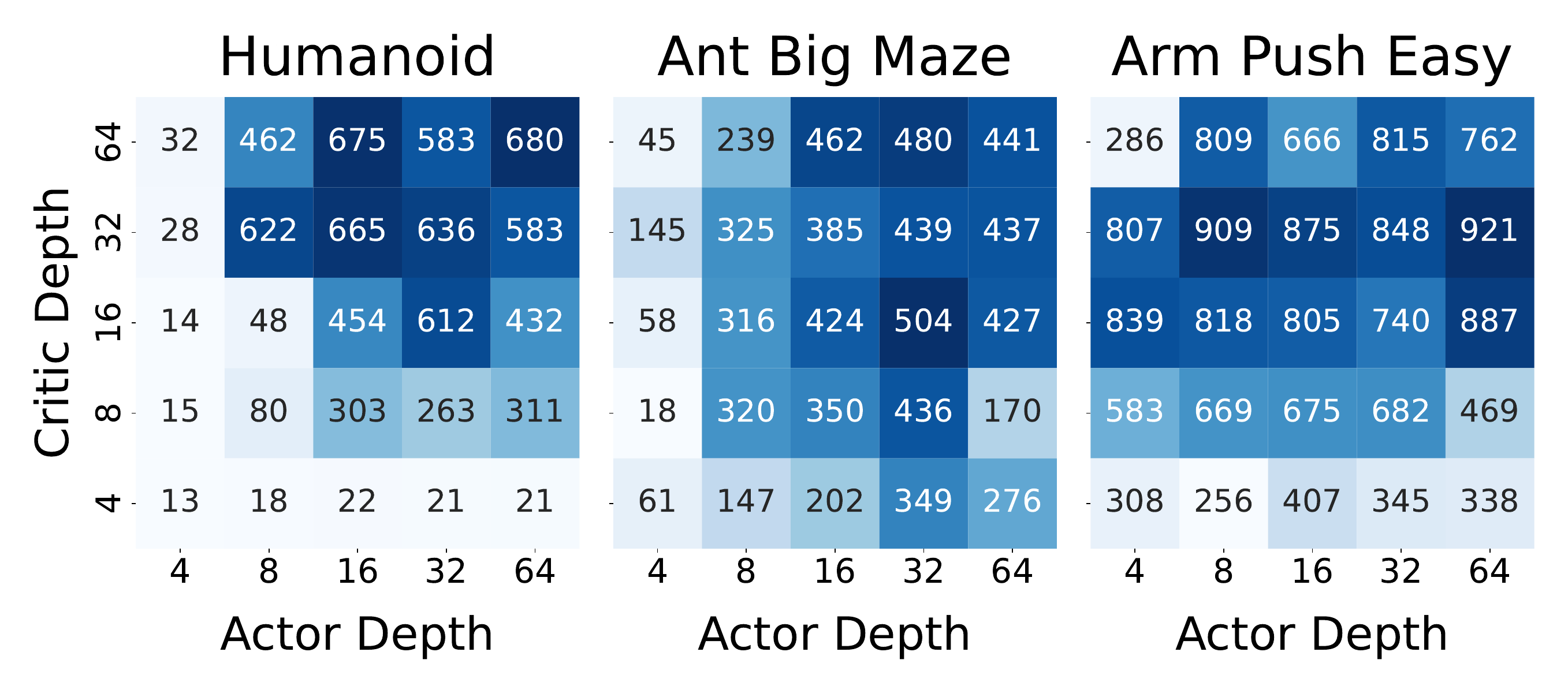}
        \vspace{-0.5em}
        \caption{\footnotesize \textbf{Actor vs. Critic.}
        In Arm Push Easy, scaling the critic is more effective; in Ant Big Maze, the actor matters more. For Humanoid, scaling both is necessary. These results suggest that actor and critic scaling can complement each other for CRL.}
        \label{fig:actor-critic}
    \end{minipage}
\end{figure}
% \begin{figure}[t]
%     \centering
%     \includegraphics[width=1\linewidth]{figures/critic-vs-actor.pdf}
%     \vspace{-1em}
%     \caption{\footnotesize \textbf{Actor vs. Critic.} In the Arm Push Easy environment, scaling the critic is more effective, whereas in the Ant Big Maze, scaling the actor has a greater impact. In the Humanoid environment, scaling both actor and critic together is essential. Overall, the results suggest that scaling both actor and critic networks can complement each other to enhance performance.}
%     \label{fig:actor-critic}
% \end{figure}

To investigate the role of scaling in the actor and critic networks, \cref{fig:actor-critic} presents the final performance for various combinations of actor and critic depths across three environments.
Prior work~\citep{naumanBiggerRegularizedOptimistic2024,leeSimBaSimplicityBias2024} focuses on scaling the critic network, finding that scaling the actor degrades performance. In contrast, while we do find that scaling the critic is more impactful in two of the three environments (Humanoid, Arm Push Easy), our method benefits from scaling the actor network jointly, with one environment (Ant Big Maze) demonstrating actor scaling to be more impactful. Thus, our method suggests that scaling both the actor and critic networks can play a complementary role in enhancing performance.

% \vspace{10pt}

\paragraph{Deep Networks Unlock Batch Size Scaling}

\begin{figure}[h]
    \centering
    % \vspace{-20pt}
    \includegraphics[width=0.8\textwidth]{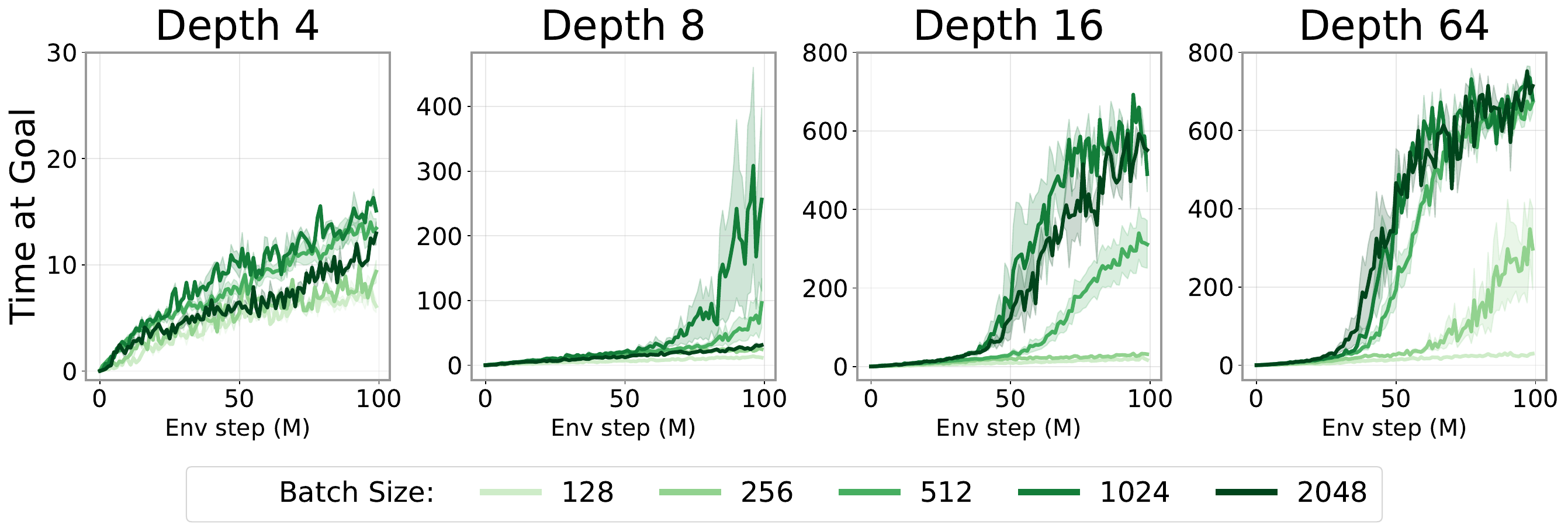}
    % \vspace{-1em}
    \caption{\footnotesize \textbf{Deeper networks unlock batch size scaling.}
    We find that as depth increases from 4 to 64 in Humanoid, larger networks can effectively leverage batch size scaling to achieve further improvements.}
    \label{fig:batch-size}
    \vspace{-10pt}
\end{figure}

Scaling batch size has been well-established in other areas of machine learning~\citep{chen2022why,zhang2024critical}. However, this approach has not translated as effectively to reinforcement learning (RL), and prior work has even reported negative impacts on value-based RL~\citep{obando2023small}. Indeed, in our experiments, simply increasing the batch size for the original CRL networks yields only marginal differences in performance (\cref{fig:batch-size}, top left).

At first glance, this might seem counterintuitive: since reinforcement learning typically involves fewer informational bits per piece of training data~\citep{lecun2016predictive}, one might expect higher variance in batch loss or gradients, suggesting the need for larger batch sizes to compensate. At the same time, this possibility hinges on whether the model in question can actually make use of a bigger batch size---in domains of ML where scaling has been successful, larger batch sizes usually bring the most benefit when coupled with sufficiently large models~\citep{zhang2024critical,chen2022why}. One hypothesis is that the small models traditionally used in RL may obscure the underlying benefits of larger batch size.

To test this hypothesis, we study the effect of increasing the batch size for networks of varying depths. As shown in \cref{fig:batch-size}, scaling the batch size becomes effective as network depth grows. This finding offers evidence that by scaling network capacity, we may simultaneously unlock the benefits of larger batch size, potentially making it an important component in the broader pursuit of scaling self-supervised RL.

% % \subsection{Does scaling depth help in the offline setting?}
% \subsection{Does Depth Scaling Improve Offline Contrastive RL?}

\paragraph{Training Contrastive RL with 1000+ Layers}
\label{sec:testing_limits}

We next study whether further increasing depth beyond 64 layers further improves performance.
% Having shown the efficacy of networks up to 64 layers deep (\cref{fig:main-figure}), we continue to examine the limits to which depth can be scaled.
We use the Humanoid maze tasks as these are both the most challenging environments in the benchmark and also seem to benefit from the deepest scaling. The results, shown in \cref{fig:scaling_limits}, indicate that performance continues to substantially improve as network depth reaches 256 and 1024 layers in the Humanoid U-Maze environment.
While we were unable to scale beyond 1024 layers due to computational constraints, we expect to see continued improvements with even greater depths, especially on the most challenging tasks.

\begin{figure}[h]
    \centering
    \includegraphics[width=1.0\linewidth]{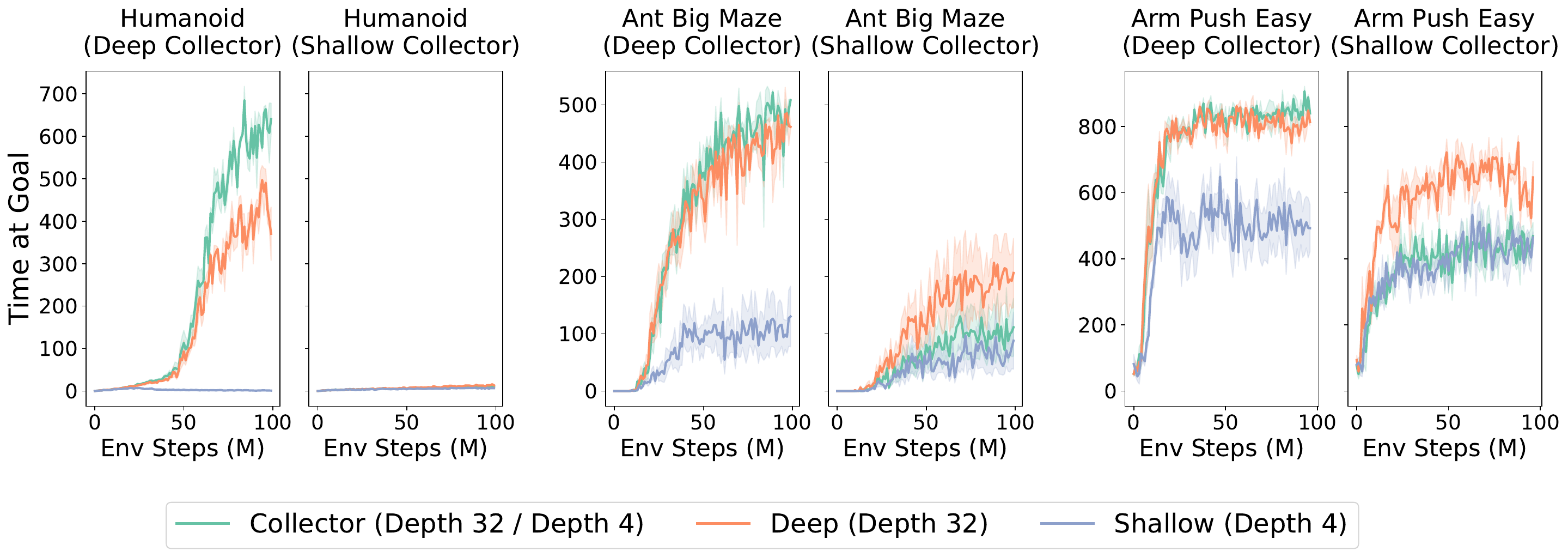}
    \vspace{-1.5em}
    \caption{\footnotesize We disentangle the effects of exploration and expressivity on depth scaling by training three networks in parallel: a “collector,” plus one deep and one shallow learner that train only from the collector’s shared replay buffer. In all three environments, when using a deep collector (i.e. good data coverage), the deep learner outperforms the shallow learner, indicating that expressivity is crucial when controlling for good exploration. With a shallow collector (poor exploration), even the deep learner cannot overcome the limitations of insufficient data coverage. As such, the benefits of depth scaling arise from a combination of improved exploration and increased expressivity working jointly.
}

    % hat even when the data the shallow and deep networks are trained on is held constant, the deep networks obtain higher success, indicating that scaling the number of layers can enhance expressivity.

    \label{fig:collector}    % \vspace{-1em}
\end{figure}

\subsection{Why Scaling Happens}
\label{sec:why_scaling_happens}

\paragraph{Depth Enhances Contrastive Representations} 

\begin{wrapfigure}{r}{0.45\textwidth}
    \centering
    \vspace{-10pt}
    \includegraphics[width=\linewidth]{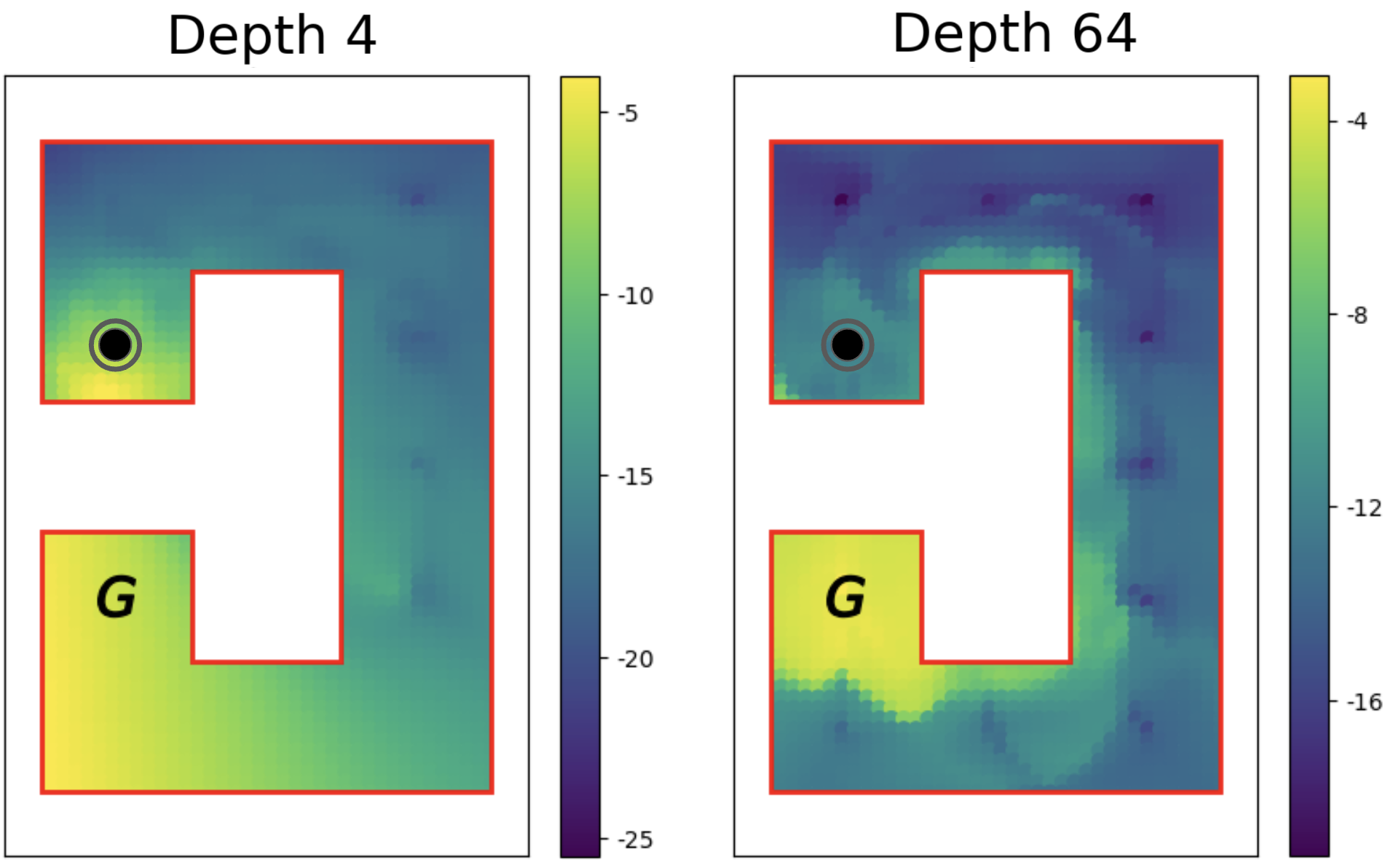}
    \vspace{-1em}
    \caption{\footnotesize \textbf{Deeper Q-functions are qualitatively different.} In the U4-Maze, the start and goal positions are indicated by the $\circledbullet$ and $\textbf{G}$ symbols respectively, and the visualized Q values are computed via the $L_2$ distance in the learned representation space, i.e., $Q(s,a,g) = \|\phi(s,a) - \psi(g)\|_2$. The shallow depth 4 network \textit{(left)} naively relies on Euclidean proximity, showing high Q values near the start despite a maze wall. In contrast, the depth 64 network \textit{(right)} clusters high Q values at the goal, gradually tapering along the interior.}
    \label{fig:Q_vis_long_horizon}
\end{wrapfigure}

The long-horizon setting has been a long-standing challenge in RL 
% \kevin {@Michal can you add a citation here},
particularly in unsupervised goal-conditioned settings where there is no auxiliary reward feedback~\citep{gupta2019relay}. The family of U-Maze environments requires a global understanding of the maze layout for effective navigation. We consider a variant of the Ant U-Maze environment, the U4-maze, in which the agent must initially move in the direction opposite the goal to loop around and ultimately reach it. As shown in \cref{fig:Q_vis_long_horizon}, we observe a qualitative difference in the behavior of the 
shallow network (depth 4) compared to the deep network (depth 64). The visualized Q-values computed from the critic encoder representations reveal that the depth 4 network seemingly relies on Euclidean distance to the goal as a proxy for the Q value, even when a wall obstructs the direct path. In contrast, the depth 64 critic network learns richer representations, enabling it to effectively capture the topology of the maze as visualized by the trail of high Q values along the inner edge. These findings suggest that increasing network depth leads to richer learned representations, enabling deeper networks to better capture environment topology and achieve more comprehensive state-space coverage in a self-supervised manner.

\paragraph{Depth Enhances Exploration and Expressivity in a Synergized Way}
Our earlier results suggested that deeper networks achieve greater state-action coverage. To better understand why scaling works, we sought to determine to whether improved data alone explains the benefits of scaling, or whether it acts in conjunction with other factors. Thus, we designed an experiment in \cref{fig:collector} in which we train three networks in parallel: one network, the “collector," interacts with the environment and writes all experience to a shared replay buffer. Alongside it, two additional "learners", one deep and one shallow, train concurrently. Crucially, these two learners never collect their own data; they train only from the collector’s buffer. This design holds the data distribution constant while varying the model’s capacity, so any performance gap between the deep and shallow learners must come from expressivity rather than exploration. When the collector is deep (e.g., depth 32), across all three environments the deep learner substantially outperforms the shallow one across all three environments, indicating that the expressivity of the deep networks is critical. On the other hand, we repeat the experiment with shallow collectors (e.g., depth 4), which explores less effectively and therefore populates the buffer with low-coverage experience. Here, both the deep and shallow learners struggle and achieve similarly poor performance, which indicates that the deep network’s additional capacity does not overcome the limitations of insufficient data coverage. As such, scaling depth enhances exploration and expressivity in a synergized way: stronger learning capacity drives more extensive exploration, and strong data coverage is essential to fully realize the power of stronger learning capacity. Both aspects jointly contribute to improved performance.

\begin{figure}[h]
    \centering
    \includegraphics[width=1.0\linewidth]{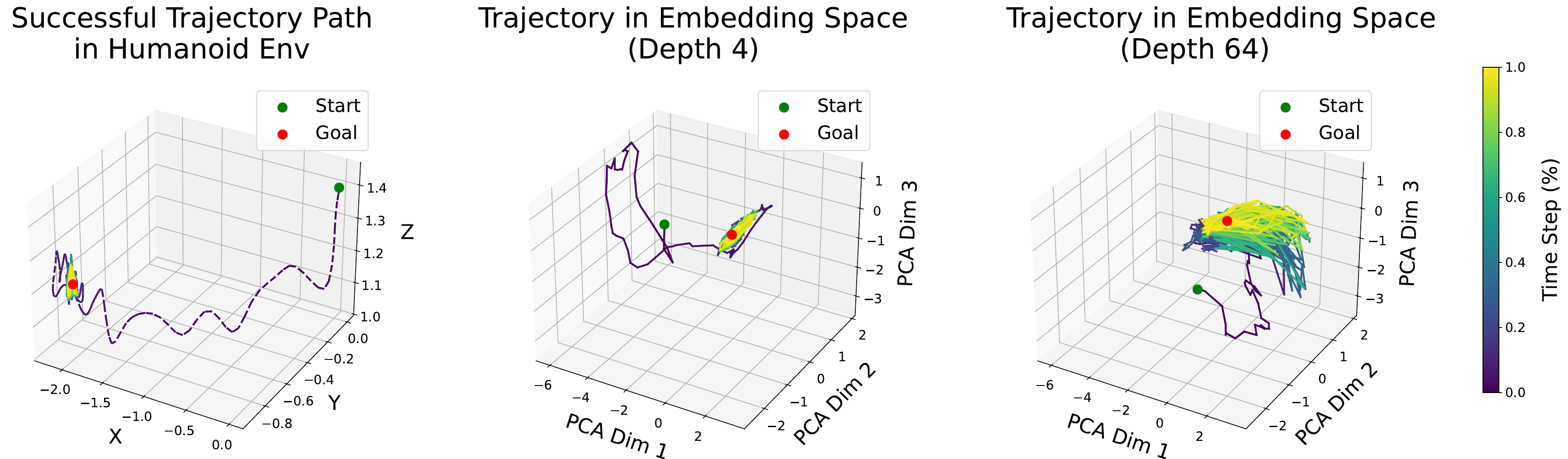}
    \vspace{+0.2em}
    \caption{\footnotesize We visualize state-action embeddings from shallow (depth 4) and deep (depth 64) networks along a successful trajectory in the Humanoid task. Near the goal, embeddings from the deep network expand across a curved surface, while those from the shallow network form a tight cluster. This suggests that deeper networks may devote greater representational capacity to regions of the state space that are more frequently visited and play a more critical role in successful task completion.}

    \label{fig:pcas}    % \vspace{-1em}
\end{figure}

\paragraph{Deep Networks Learn to Allocate Greater Representational Capacity to States Near the Goal}

In \Cref{fig:pcas} we take a successful trajectory in the Humanoid environment and visualize the embeddings of state-action encoder along this trajectory for both deep vs. shallow networks. While the shallow network (Depth 4) tends to cluster near-goal states tightly together, the deep network produces more "spread out" representations. This distinction is important: in a self-supervised setting, we want our representations to separate states that matter—particularly future or goal-relevant states—from random ones. As such, we want to allocate more representational capacity to such critical regions. This suggests that deep networks may learn to allocate representational capacity more effectively to state regions that matter most for the downstream task.

% \subsection{Increasing Depth Results in (Some) Stitching} 
\paragraph{Deeper Networks Enable Partial Experience Stitching} 

\begin{wrapfigure}{r}{0.5\textwidth}
    \centering
    \vspace{-1em}
    \includegraphics[width=0.99\linewidth]{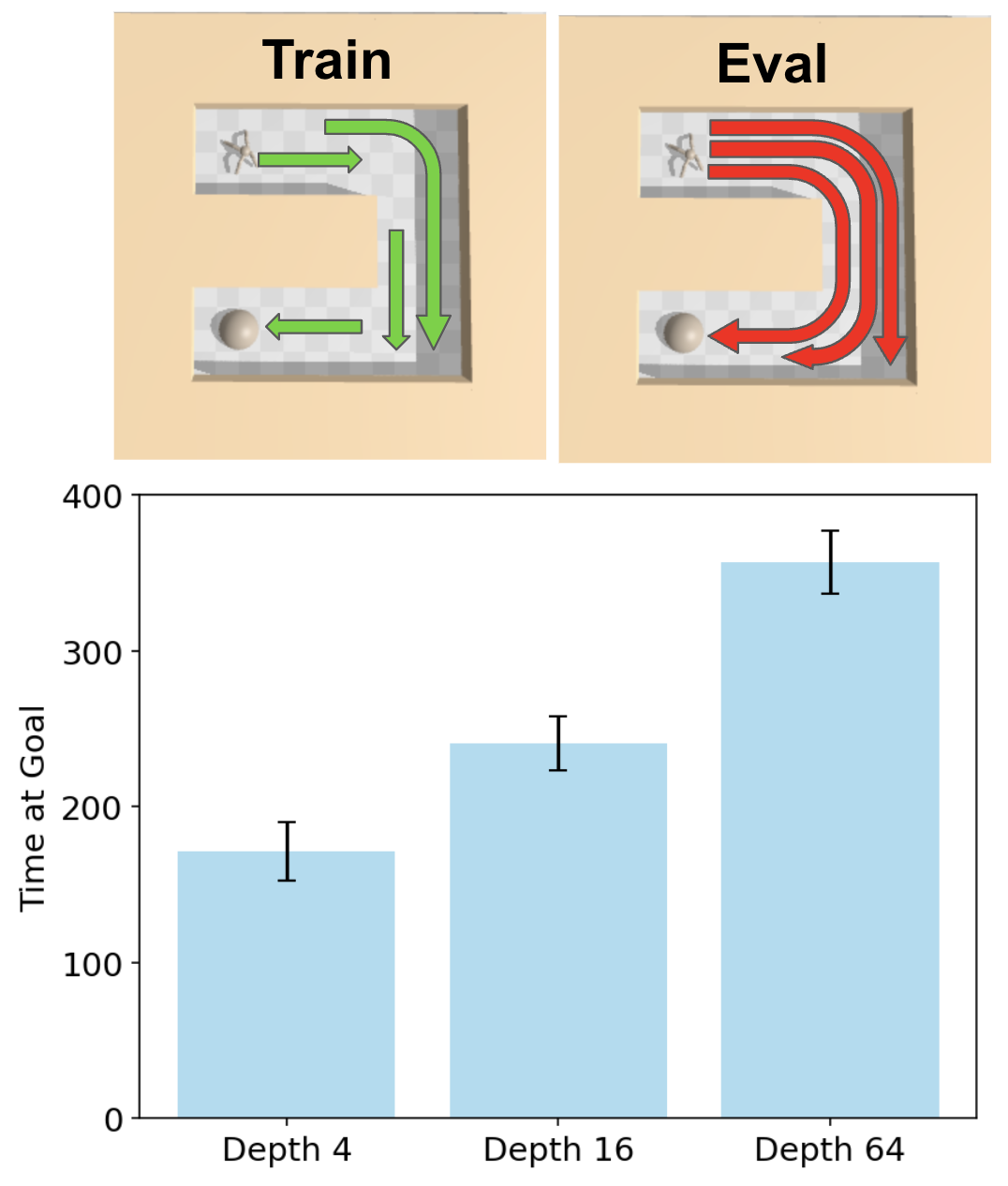}
    \vspace{-1em}
    \caption{\footnotesize \textbf{Deeper networks exhibit improved generalization.} \textit{(Top left)} We modify the training setup of the Ant U-Maze environment such that start-goal pairs are separated by $\leq3$ units. This design guarantees that no evaluation pairs \textit{(Top right)} were encountered during training, testing the ability for combinatorial generalization via stitching. \textit{(Bottom)} Generalization ability improves as network depth grows from 4 to 16 to 64 layers.}
    \label{fig:generalization}
    \vspace{-18pt}
\end{wrapfigure}

Another key challenge in reinforcement learning is learning policies that can generalize to tasks unseen during training. To evaluate this setting, we designed a modified version of the Ant U-Maze environment. As shown in \cref{fig:generalization} (top right), the original JaxGCRL benchmark assesses the agent's performance on the three farthest goal positions located on the opposite side of the wall. However, instead of training on all possible subgoals (a superset of the evaluation state-goal pairs), we modified the setup to train on start-goal pairs that are at most 3 units apart, ensuring that none of the evaluation pairs ever appear in the training set. \cref{fig:generalization} demonstrates that depth 4 networks show limited generalization, solving only the easiest goal (4 units away from the start). Depth 16 networks achieve moderate success, while depth 64 networks excel, sometimes solving the most challenging goal position. These results suggest that the increasing network depth results in some degree of stitching, combining $\leq$3-unit pairs to navigate the 6-unit span of the U-Maze. 
% These results demonstrate that scaling depth can help learn improved generalization ability and robustness.
% as depth scales, one can yield improved generalization properties.

\paragraph{The (CRL) Algorithm is Key}

In Appendix~\ref{sec:add_baselines}, we show that scaled CRL outperforms other baseline goal-conditioned algorithms and advance the SOTA for goal-conditioned RL. We observe that for temporal difference methods (SAC, SAC+HER, TD3+HER), the performance saturates for networks of depth 4, and there is either zero or negative performance gains from deeper networks. This is in line with previous research showing that these methods benefit mainly from width~\citep{leeSimBaSimplicityBias2024,naumanBiggerRegularizedOptimistic2024}. These results suggest that the self-supervised CRL algorithm is critical.

We also experiment with scaling more self-supervised algorithms, namely Goal-Conditioned Behavioral Cloning (GCBC) and Goal-Conditioned Supervised Learning (GCSL). While these methods yield zero success in certain environments, they show some utility in arm manipulation tasks. Interestingly, even a very simple self-supervised algorithm like GCBC benefits from increased depth. This points to a promising direction for future work of further investigating other self-supervised methods to uncover potentially different or complementary recipes for scaling self-supervised RL.

Finally, recent work has augmented goal-conditioned RL with quasimetric architectures, leveraging the fact that temporal distances satisfy a triangle inequality–based invariance. In Appendix~\ref{sec:add_baselines}, we also investigate whether the depth scaling effect persists when applied to these quasimetric networks.

% \begin{figure}[t]
%     \centering
%     \includegraphics[width=\linewidth]{figures/depth-scaling.pdf}
%     \vspace{-1em}
%     \caption{\footnotesize \textbf{Testing the limits of scale.}
%     We extend the results from \cref{fig:main-figure} by scaling networks even further on the challenging Humanoid maze environments. We observe continued performance improvements with network depths of 256 and 1024 layers on Humanoid U-Maze. Note that for the 1024-layer networks, we observed the actor loss exploding at the onset of training, so we maintained the actor depth at 512 while using 1024-layer networks only for the two critic encoders.
%     }
%     \label{fig:scaling_limits}
%     % \vspace{-1em}
% \end{figure}
\begin{wrapfigure}{r}{0.5\textwidth}
    \centering
    \vspace{-1.25em}
    \includegraphics[width=0.48\textwidth]{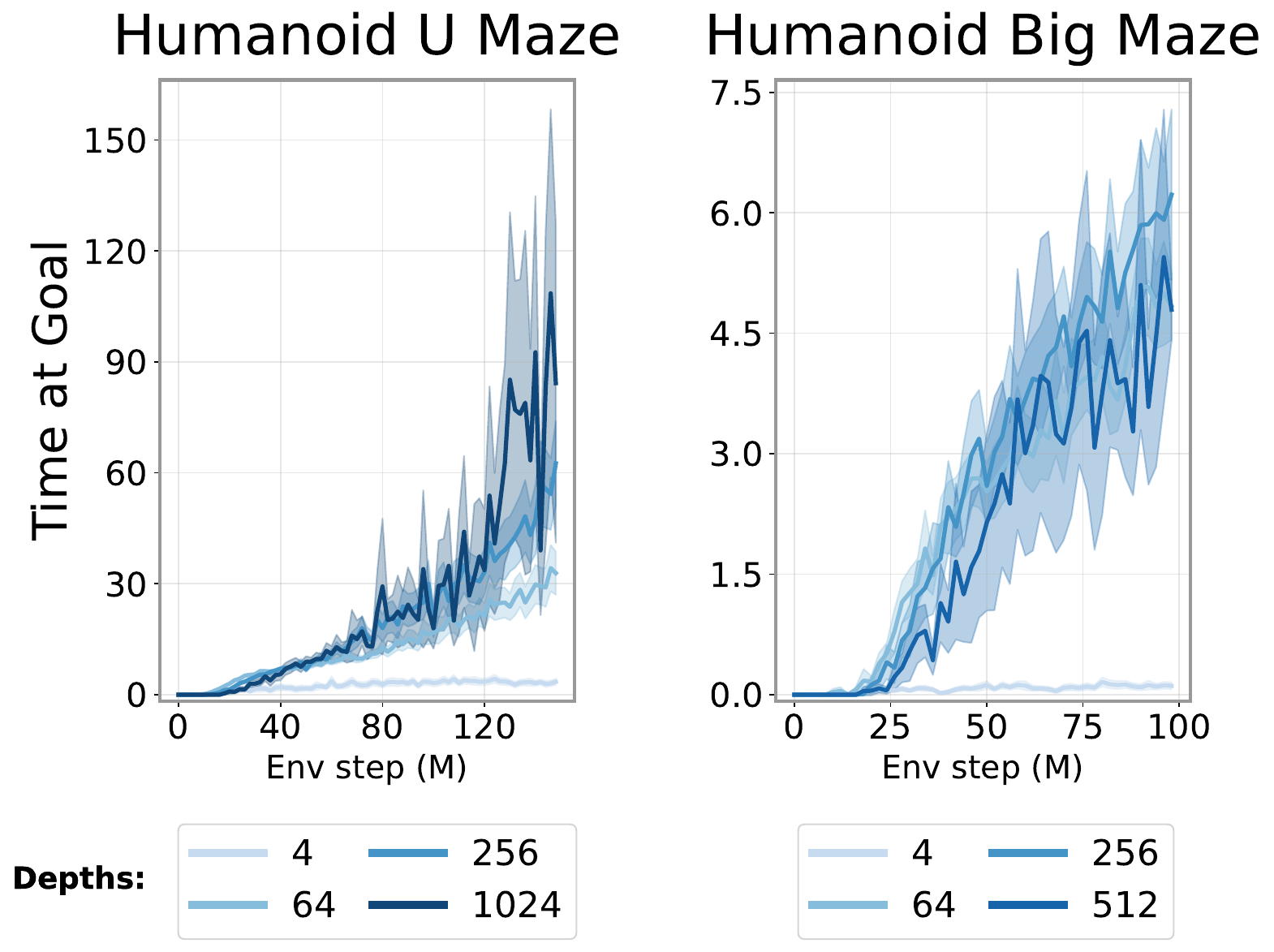}
    \vspace{-0.25em}
    \caption{\footnotesize \textbf{Testing the limits of scale.}
    We extend the results from \cref{fig:main-figure} by scaling networks even further on the challenging Humanoid maze environments. We observe continued performance improvements with network depths of 256 and 1024 layers on Humanoid U-Maze. Note that for the 1024-layer networks, we observed the actor loss exploding at the onset of training, so we maintained the actor depth at 512 while using 1024-layer networks only for the two critic encoders.}
    \label{fig:scaling_limits}
    \vspace{-35pt}
\end{wrapfigure}

% \begin{figure}[t] \footnotesize 
% \begin{tcolorbox}[colback=white, colframe=black, boxrule=1pt, left=5pt, right=5pt, top=5pt, bottom=5pt]
% \textbf{Summary of key empirical findings}:
%     \begin{itemize}\setlength{\itemsep}{0pt}\addtolength{\leftskip}{-5pt}
%         \item CRL is scalable to depths unattainable by other RL proprioceptive algorithms (1000+ layers), perhaps due to its self-supervised nature.
%         \item Both width and depth are key factors influencing CRL's performance, but depth achieves greater performance and better parameter-efficiency (similar performance for $50\times$ smaller models).
%         % \item \textbf{Depth is the primary factor driving the performance. In contrast, width scaling yields marginal benefits while being computationally more expensive.}
%         \item We observe signs of emergent behaviors in CRL with deep neural networks, such as humanoid learning to walk and navigate a maze.
%         \item Scale unlocks learning difficult maze topologies.
%         \item Batch size scaling occurs in CRL for deep networks. 
%         \item CRL benefits from both the actor and critic scale.
%     \end{itemize}
% \end{tcolorbox}
% \end{figure}

\subsection{Does Depth Scaling Improve Offline Contrastive RL?}
In preliminary experiments, we evaluated depth scaling in the offline goal-conditioned setting using OGBench ~\citep{park2024ogbench}. We found little evidence that increasing the network depth of CRL improves performance in this offline setting. 
To further investigate this, we conducted ablations: \textit{(1)} scaling critic depth while holding the actor at 4 or 8 layers, and \textit{(2)} applying cold initialization to the final layers of the critic encoders~\citep{zheng2024stabilizing}. In all cases, baseline depth 4 networks often had the highest success. A key direction for future work is to see if our method can be adapted to enable scaling in the offline setting.

\section{Conclusion}
% \ben{Speculative sentences for use at the end of the introduction or conclusion.}

% \kevin {@Ben we thought the paper might be strongest if we leave you the creative freedom to write the conclusion here, let us know your thoughts on this :) We noticed that you included a lot of really good speculative sentences/ideas when making the initial pass through the paper. Currently via the intro and RW, we've instantiated a very specific framing for the paper which follows the thought-lines that we've come up with after thinking deeply. If we were to write a conclusion, it would likely come off as sort of summarizing/rehashing the same themes. 

% On the other hand, I feel like you'd be a very good person to come in at the end and infuse some fresh and new insights/thoughts/perspectives on the high-level with regard to the implications/perspective of our work in the context for RL (i.e. stuff like "RL often reduces problems to language/vision, when can RL train its own big models", or "RL methods inherently address this by jointly optimizing both the model and the data collection process through exploration, and scaling it up can yield fruits unattainable in other current paradigms" -- just examples for illustrations, don't have to use them). Could you help to draft a conclusion that ties everything together while introducing your creative ideas?}

Arguably, much of the success of vision and language models today is due to the emergent capabilities they exhibit from scale~\citep{srivastava2023beyond}, leading to many systems reducing the RL problem to a vision or language problem. A critical question for large AI models is: where does the data come from? Unlike supervised learning paradigms, RL methods inherently address this by jointly optimizing both the model and the data collection process through exploration. 
Ultimately, determining effective ways of building RL systems that demonstrate emergent capabilities may be important for transforming the field into one that trains its own large models. 
We believe that our work is a step towards these systems. By integrating key components for scaling up RL into a single approach, we show that model performance consistently improves as scale increases in complex tasks. In addition, deep models exhibit qualitatively better behaviors which might be interpreted as implicitly acquired skills necessary to reach the goal.

% Highlight that for big AI models today, the big elephant in the room is data. Where does these data come from? RL methods are jointly optimizing data and the model. They tell you how to collect data (i.e., exploration). 

\paragraph{Limitations.}
The primary limitations of our results are that scaling network depth comes at the cost of compute. An important direction for future work is to study how distributed training might be used to leverage even more compute, and how techniques such as pruning and distillation might be used to decrease the computational costs.

\paragraph{Impact Statement} This paper presents work whose goal is to advance the field of Machine Learning. There are many potential societal consequences of our work, none which we feel must be specifically highlighted here.

\paragraph{Acknowledgments.}
We gratefully acknowledge Nathaniel Chen, Galen Collier, and the full staff of Princeton Research Computing for their invaluable assistance. We also thank Colin Lu for his discussions and contributions to this work. 
This research was also partially supported by the National Science Centre, Poland (grant no. 2023/51/D/ST6/01609); the Princeton Laboratory for Artificial Intelligence under Award 2025-97; and the Warsaw University of Technology through the Excellence Initiative: Research University (IDUB) program. Finally, we would also like to thank Jens Tuyls and Harshit Sikchi for providing helpful commends and feedback on the manuscript.

\clearpage
% \thebib
{\footnotesize
\bibliographystyle{apalike}
\bibliography{references}
}

\newpage
\appendix
\onecolumn

% \section*{Appendix: Supplementary Overview}
% \addcontentsline{toc}{section}{Appendix: Supplementary Overview}

\section{Additional Experiments}

\label{sec:add_baselines}

\addtocounter{figure}{-1}

\subsection{Scaled CRL Outperforms All Other Baselines on 8 out of 10 Environments}

\begin{figure}[h]
    \centering
    \includegraphics[width=0.95\linewidth]{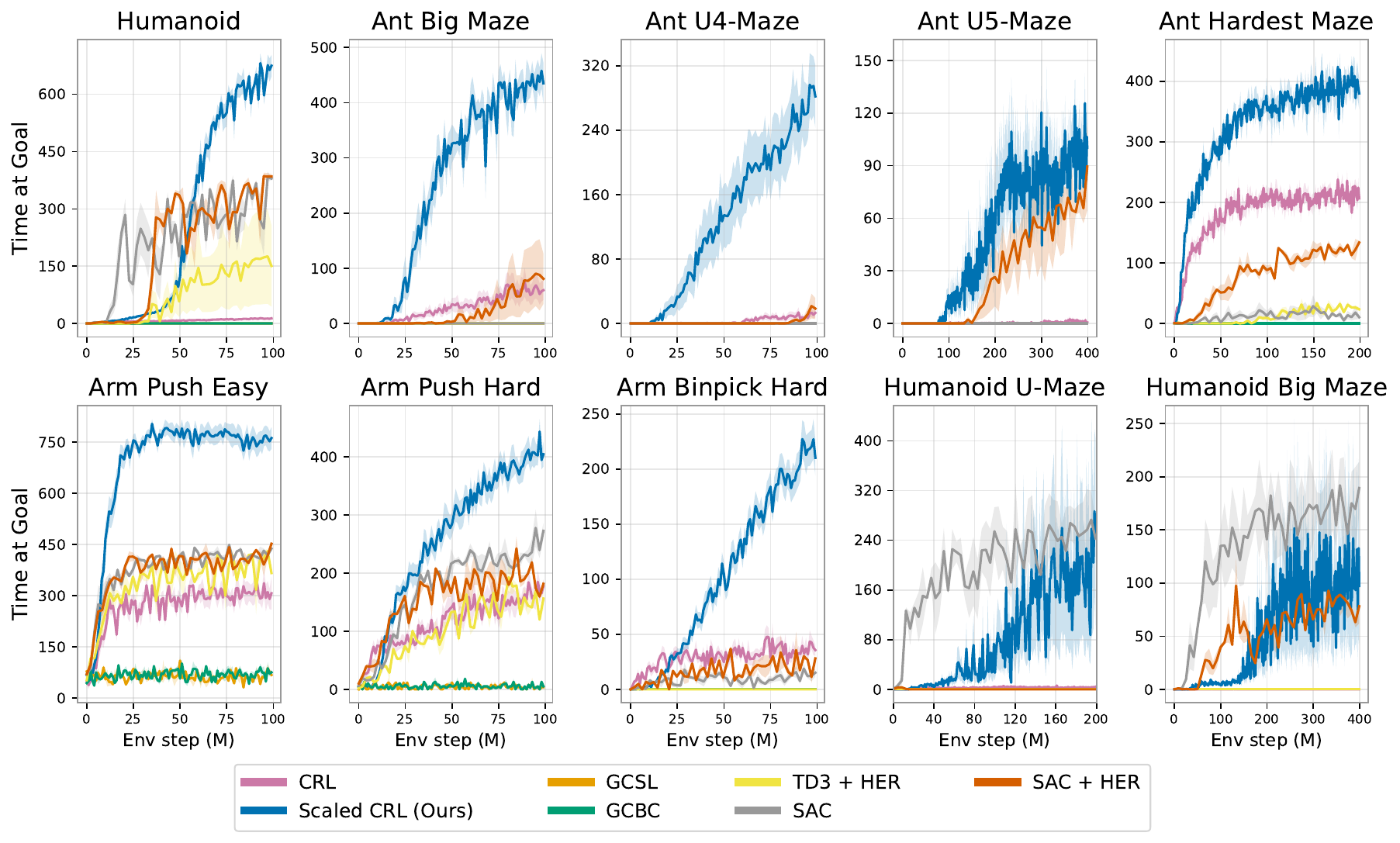}
    \vspace{-0.5em}
    \caption{Scaled CRL (Ours) outperforms baselines CRL (original), SAC, SAC+HER, TD3+HER, GCSL, and GCBC in 8 out 10 environments.}

    \label{fig:baselines}    % \vspace{-1em}
\end{figure}

In Figure \ref{fig:main-figure}, we demonstrated that increasing the depth of the CRL algorithm leads to significant performance improvements over the original CRL (see also Table \ref{tab:performance_metrics}). Here, we show that these gains translate to state-of-the-art results in online goal-conditioned RL, with Scaled CRL outperforming both standard TD-based methods such as SAC, SAC+HER, and TD3+HER, as well as self-supervised imitation-based approaches like GCBC and GCSL.

\subsection{The CRL Algorithm is Key: Depth Scaling is Not Effective on Other Baselines}

\begin{figure}[h]
    \centering
    \includegraphics[width=0.905\linewidth]{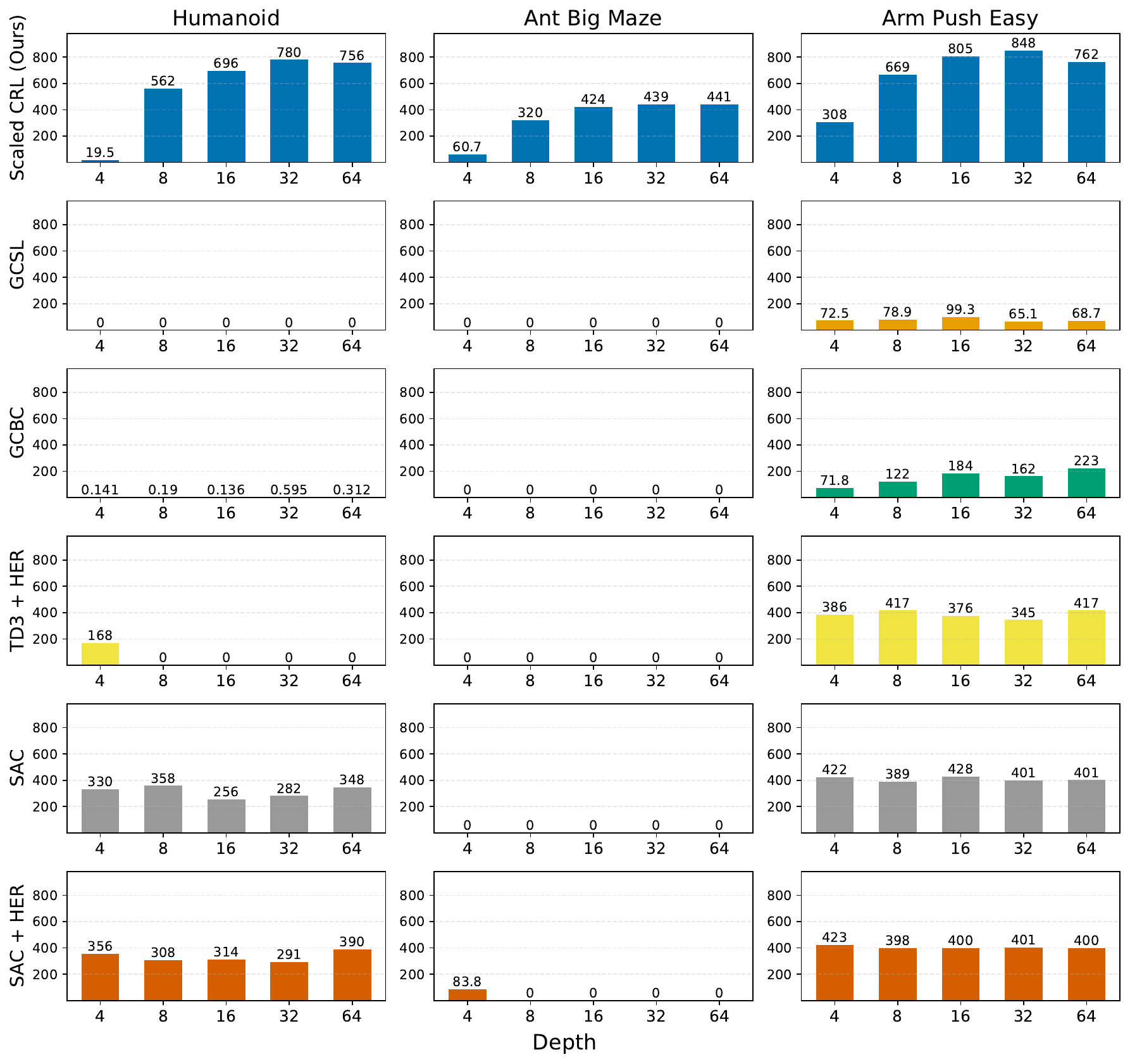}
    % \vspace{-1em}
    \caption{Depth scaling yields limited gains for SAC, SAC+HER, TD3+HER, GCSL, and GCBC.}

    \label{fig:baselines_scaling}    % \vspace{-1em}
\end{figure}

Next, we investigate whether increasing network depth in the baseline algorithms yields similar performance improvements as observed in CRL. We find that SAC, SAC+HER, and TD3+HER do not benefit from depths beyond four layers, which is consistent with prior findings~\citep{leeSimBaSimplicityBias2024,naumanBiggerRegularizedOptimistic2024}. Additionally, GCSL and GCBC fail to achieve any meaningful performance on the Humanoid and Ant Big Maze tasks. Interestingly, we do observe one exception, as GCBC exhibits improved performance with increased depth in the Arm Push Easy environment.

\begin{table}[h]
\caption{
% Success of depth-4 and depth-64 networks across 10 tasks. As observation dimension grows, the performance increase from depth scaling grows even greater.
Increasing network depth (depth $D = 4 \rightarrow 64$) increases performance on CRL (\cref{fig:main-figure}). Scaling depth exhibits the greatest benefits on tasks with the largest observation dimension (\textit{Dim}).
}
\centering
\small
\renewcommand{\arraystretch}{1.09}
\setlength{\tabcolsep}{4.5pt} % Adjust column spacing
\begin{tabular}{lccccc}
\toprule
\textbf{Task} & \textbf{Dim} & 
\textbf{\textit{$\textbf{D = }$}4} 
& 
\textbf{\textit{$\textbf{D = }$}64} 
& \textbf{Imprv.} \\
\midrule
Arm Binpick Hard & \multirow{3}{*}{\textit{17}} & $38$ {\tiny$\mathbf{\pm} 4$} & $219$ {\tiny$\mathbf{\pm} 15$}  & $5.7 \times$ \\
Arm Push Easy    &  & $308$ {\tiny$\mathbf{\pm} 33$}  & $762$ {\tiny$\mathbf{\pm} 30$}  & $2.5 \times$  \\
Arm Push Hard    &  & $171$ {\tiny$\mathbf{\pm} 11$}  & $410$ {\tiny$\mathbf{\pm} 13$}  & $2.4 \times$  \\
\cline{1-5}
\addlinespace[2pt] % Adds 5pt vertical space
Ant U4-Maze      & \multirow{4}{*}{\textit{29}} & $11.4$ {\tiny$\mathbf{\pm } 4.1$} & $286$ {\tiny$\mathbf{\pm} 36$}  & $25 \times$  \\
Ant U5-Maze      &  & $0.97$ {\tiny$\mathbf{\pm} 0.7$} & $61$ {\tiny$\mathbf{\pm} 18$}  & $63 \times$  \\
Ant Big Maze     &  & $61$ {\tiny$\mathbf{\pm} 20$} & $441$ {\tiny$\mathbf{\pm} 25$}  & $7.3 \times$  \\
Ant Hardest Maze &  & $215$ {\tiny$\mathbf{\pm} 8$} & $387$ {\tiny$\mathbf{\pm} 21$}  & $1.8 \times$  \\
\cline{1-5}
\addlinespace[2pt] % Adds 5pt vertical space
Humanoid         & \multirow{3}{*}{\textit{268}} & 
$12.6$ {\tiny$\mathbf{\pm} 1.3$} & $649$ {\tiny$\mathbf{\pm} 19$}  & $52 \times$  \\
Humanoid U-Maze  &  & $3.2$ {\tiny$\mathbf{\pm} 1.2$} & $159$ {\tiny$\mathbf{\pm} 33$}  & $50 \times$  \\
Humanoid Big Maze &  & $0.06$ {\tiny$\mathbf{\pm} 0.04$} & $59$ {\tiny$\mathbf{\pm} 21$} & $1051 \times$  \\
\bottomrule

\end{tabular}

\label{tab:performance_metrics}
\end{table}

\subsection{Additional Scaling Experiments: Offline GCBC, BC, and QRL}

\begin{figure}[h]
    \centering
    \includegraphics[width=.99\linewidth]{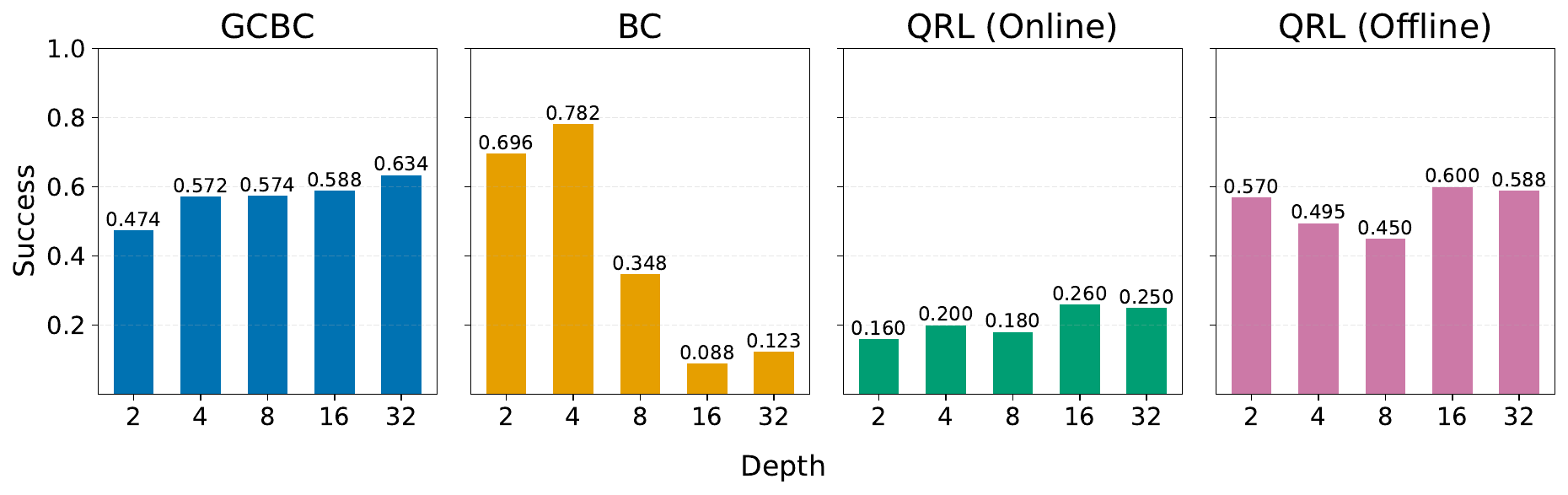}
    \caption{
    Our approach successfully scales depth in offline GCBC on \textit{antmaze-medium-stitch} (OGBench). 
    In contrast, scaling depth for BC (\textit{antmaze-giant-navigate}, expert SAC data) and for both online (\textit{FetchPush}) and offline QRL (\textit{pointmaze-giant-stitch}, OGBench) yield negative results.
    }
    \label{fig:ogbench}
\end{figure}

We further investigate several additional scaling experiments. As shown in \cref{fig:ogbench}, our approach successfully scales with depth in the offline GCBC setting on the \textit{antmaze-medium-stitch} task from OGBench. We find that our the combination of layer normalization, residual connections, and Swish activations is critical, suggesting that our architectural choices may be applied to unlock depth scaling in other algorithms and settings. We also attempt to scale depth for behavioral cloning and the QRL~\citep{wang2023optimalgoalreachingreinforcementlearning} algorithm---in both of these cases, however, we observe negative results.

\subsection{Can Depth Scaling also be Effective for Quasimetric Architectures?}

Prior work~\citep{DBLP:conf/icml/00010IZ23,liuMetricResidualNetworks2023} has found that temporal distances satisfy an important invariance property, suggesting the use of quasimetric architectures when learning temporal distances. Our next experiment tests whether changing the architecture affects the scaling properties of self-supervised RL. Specifically, we use the CMD-1 algorithm~\citep{myers2024learning}, which employs a backward NCE loss with MRN representations. The results indicate that scaling benefits are not limited to a single neural network parametrization. However, MRN’s poor performance on the Ant U5-Maze task suggests further innovation is needed for consistent scaling with quasimetric models.

\begin{figure}[h]
    \centering
    \includegraphics[width=0.95\linewidth]{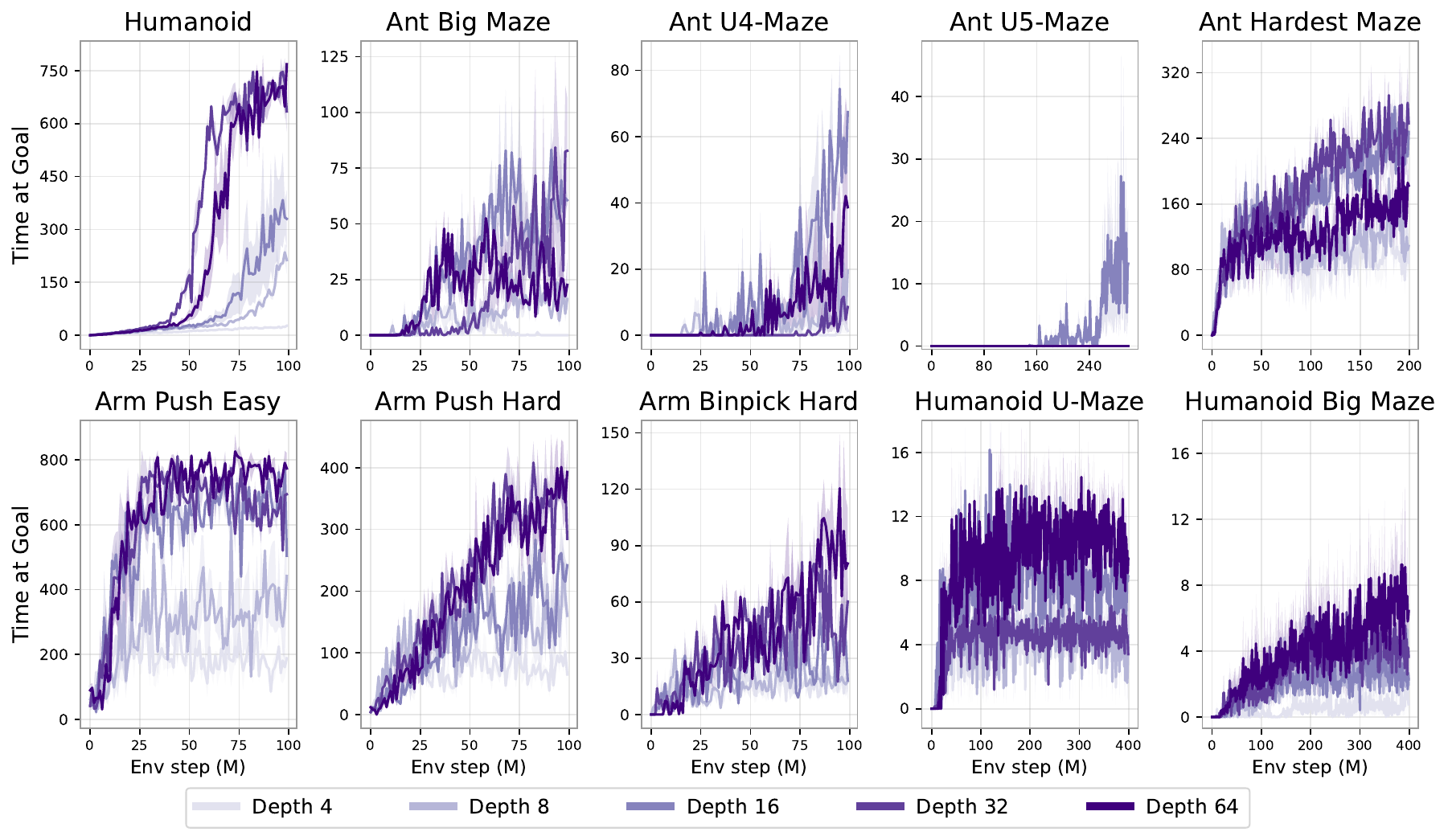}
    \vspace{-0.25em}
    \caption{Performance of depth scaling on CRL augmented with quasimetric architectures (CMD-1).}

    \label{fig:mrn}    % \vspace{-1em}
\end{figure}

\subsection{Additional Architectural Ablations: Layer Norm and Swish Activation}

\begin{figure}[h]
    \centering
    \includegraphics[width=.95\linewidth]{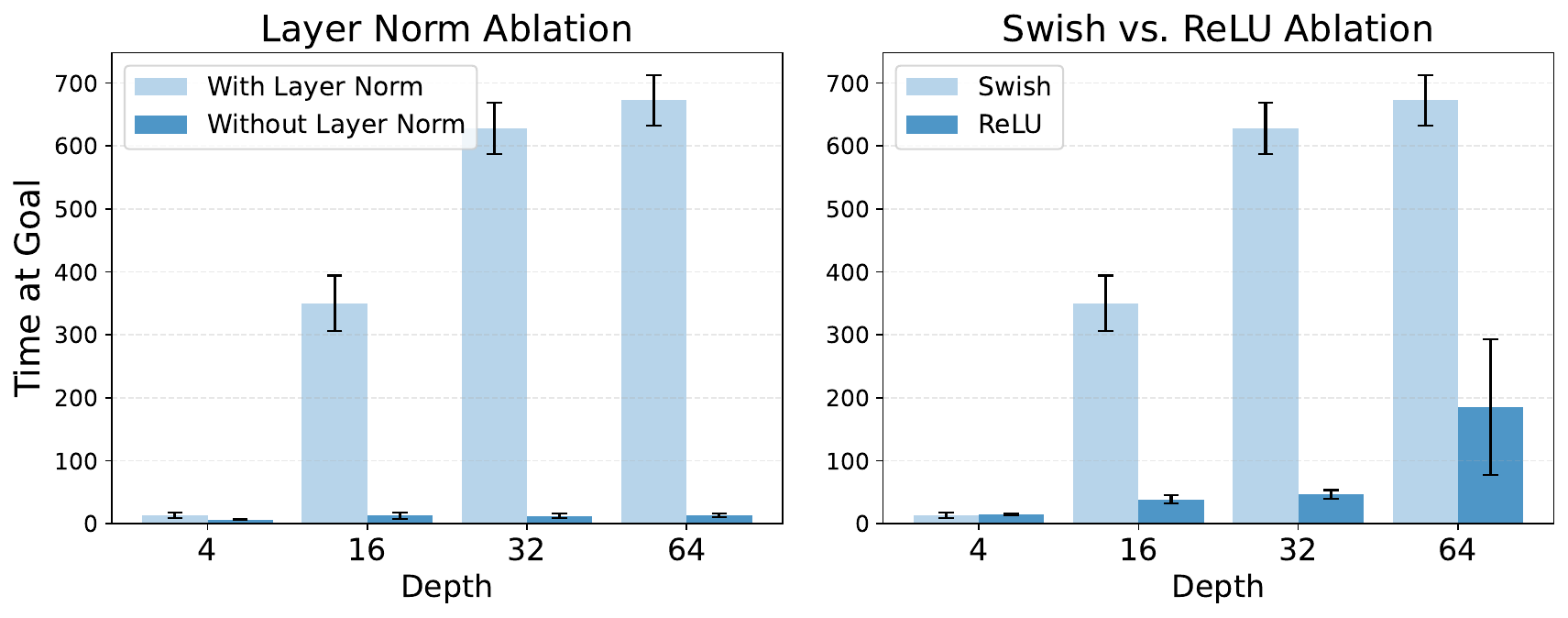}
    \caption{(Left) Layer Norm is essential for scaling depth. (Right) Scaling with ReLU activations leads to worse performance compared to Swish activations.}
    \label{fig:arch_ablations}
\end{figure}

We conduct ablation experiments to validate the architectural choices of layer norm and swish activation. \cref{fig:arch_ablations} shows that removing layer normalization performs significantly worse. Additionally, scaling with ReLU significantly hampers scalability. These results, along with \cref{fig:shallow-networks} show that all of our architectural components—residual connections, layer norm, and swish activations—are jointly essential to unlocking the full performance of depth scaling.

\subsection{Can We Integrate Novel Architectural Innovations from the Emerging RL Scaling Literature?}

Recently, Simba-v2 proposed a new architecture for scalable RL. Its key innovation is the replacement of layer normalization with hyperspherical normalization, which projects network weights onto the unit-norm hypersphere after each gradient update. As shown, the same depth-scaling trends hold when adding hyperspherical normalization to our architecture, and it further improves the sample efficiency of depth scaling. This demonstrates that our method can naturally incorporate new architectural innovations emerging in the RL scaling literature.

\begin{table}[htbp]
\centering
\caption{Integrating hyperspherical normalization in our architecture enhances the sample efficiency of depth scaling.}
\label{tab:hyperspherical_normalization}
\begin{tabular}{ccc}
% -------- Subtable 1 --------
\begin{tabular}{c|ccc}
\multicolumn{4}{c}{\textbf{Steps to reach $\geq$200 success}} \\ 
\toprule
\textbf{Depth} & 4 & 16 & 32 \\ \midrule
\textbf{With} & -- & \textbf{50} & \textbf{42} \\
\textbf{Without} & -- & 64 & 54 \\
\bottomrule
\end{tabular}
&
% -------- Subtable 2 --------
\begin{tabular}{c|ccc}
\multicolumn{4}{c}{\textbf{Steps to reach $\geq$400 success}} \\ 
\toprule
\textbf{Depth} & 4 & 16 & 32 \\ \midrule
\textbf{With} & -- & \textbf{62} & \textbf{48} \\
\textbf{Without} & -- & 75 & 64 \\
\bottomrule
\end{tabular}
&
% -------- Subtable 3 --------
\begin{tabular}{c|ccc}
\multicolumn{4}{c}{\textbf{Steps to reach $\geq$600 success}} \\ 
\toprule
\textbf{Depth} & 4 & 16 & 32 \\ \midrule
\textbf{With} & -- & \textbf{77} & \textbf{67} \\
\textbf{Without} & -- & -- & 77 \\
\bottomrule
\end{tabular}
\\
\end{tabular}

\end{table}

\subsection{Residuals Norms in Deep Networks}

% Prior work has observed that the norm of residuals in deeper network layers tends to be smaller than in earlier layers [cite]. We investigate whether this pattern also holds in our setting. For critic networks, we find that this trend generally persists, especially in very deep architectures (e.g., depth 256). This effect is not as apparent in actor networks.

% Prior work finds smaller residual norms in deeper layers [cite], a trend we also observe for our critic networks, especially in very deep networks (e.g., depth 256). Not as much in actor.

Prior work has noted decreasing residual activation norms in deeper layers~\citep{DBLP:conf/iclr/ChangMHTB18}. We investigate whether this pattern also holds in our setting. For the critic, the trend is generally evident, especially in very deep architectures (e.g., depth 256). The effect is not as pronounced in the actor.

\begin{figure}[h]
    \centering
    \vspace{-0.1em}
    \includegraphics[width=0.75\linewidth]{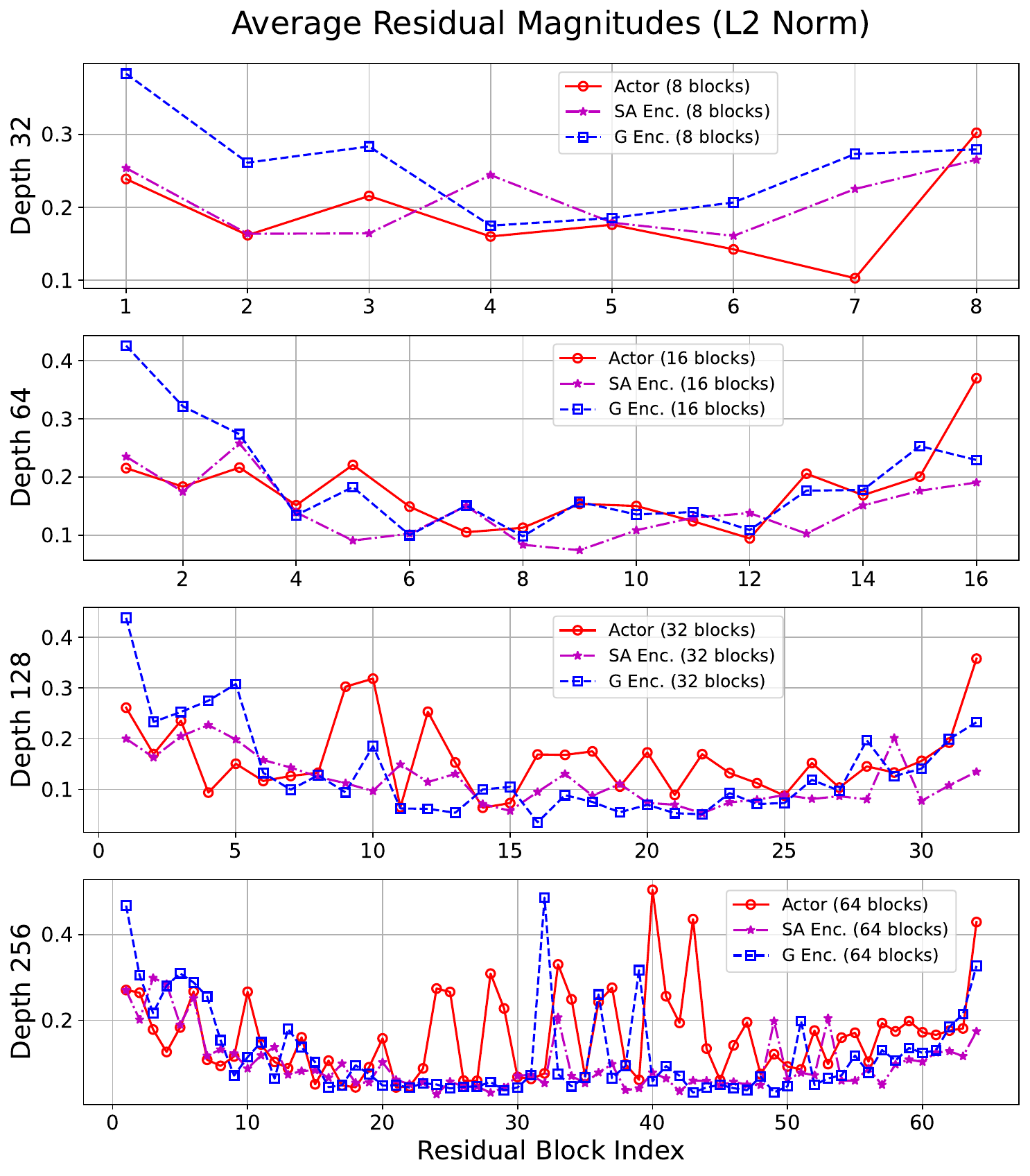}
    % \vspace{0.5em}
    \vspace{-0.65em}
    \caption{L2 norms of residual activations in networks with depths of 32, 64, 128, and 256.}

    \label{fig:mrn}    % \vspace{-1em}
\end{figure}

\subsection{Scaling Depth for Offline Goal-conditioned RL}

\vspace{1em}

\begin{figure}[h]
    \centering
    \includegraphics[width=0.95\linewidth]{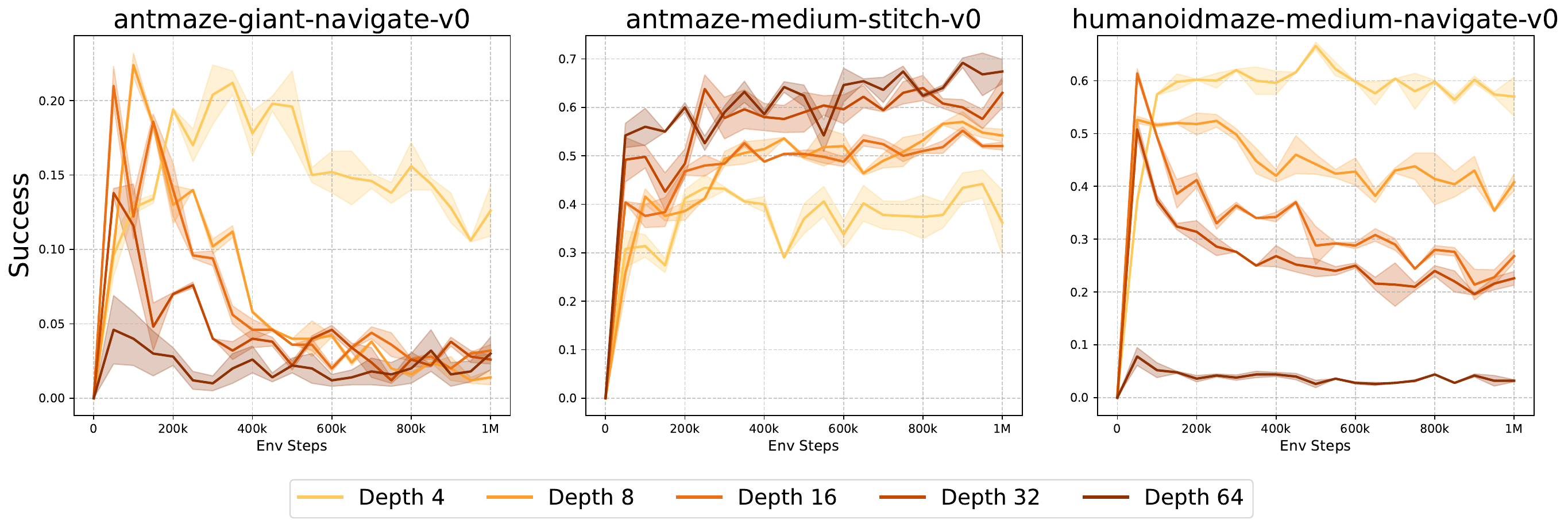}
    \vspace{-0.5em}
    \caption{To evaluate the scalability of our method in the offline setting, we scaled model depth on OGBench~\citep{park2024ogbench}. In two out of three environments, performance drastically declined as depth scaled from 4 to 64, while a slight improvement was seen on antmaze-medium-stitch-v0. Successfully adapting our method to scale offline GCRL is an important direction for future work.}

    \label{fig:offline}    % \vspace{-1em}
\end{figure}

\section{Experimental Details}
\label{app:technical_details}

\subsection{Environment Setup and Hyperparameters}

\begin{figure}[h]
    \centering
    \includegraphics[width=0.7\linewidth]{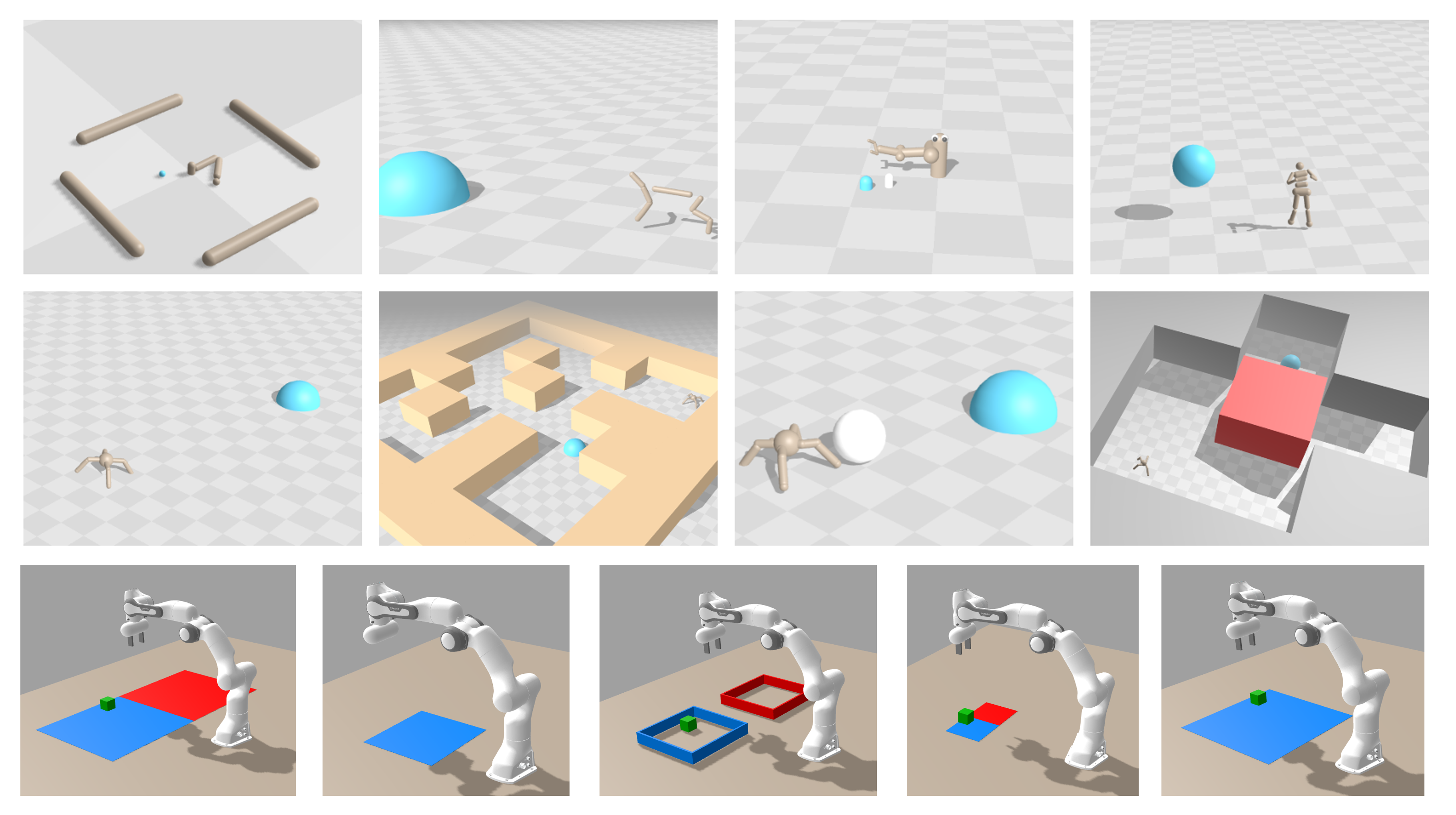}
    \caption{The scaling results of this paper are demonstrated on the JaxGCRL benchmark, showing that they replicate across a diverse range of locomotion, navigation, and manipulation tasks. These tasks are set in the online goal-conditioned setting where there are no auxiliary rewards or demonstrations. Figure taken from~\citep{bortkiewicz2024accelerating}.}
%         % \textbf{The JaxGCRL benchmark} spans a suite of locomotion, navigation, and manipulation tasks in the goal-conditioned setting with no auxiliary rewards or demonstrations. 
%     % Figure from~\citep{bortkiewicz2024accelerating}.
    \label{fig:pictures}
\end{figure}

Our experiments use the JaxGCRL suite of GPU-accelerated environments, visualized in \cref{fig:pictures}, and a contrastive RL algorithm with hyperparameters reported in \cref{table:params}. In particular, we use 10 environments, namely: \texttt{ant\_big\_maze, ant\_hardest\_maze, arm\_binpick\_hard, arm\_push\_easy, arm\_push\_hard, humanoid, humanoid\_big\_maze, humanoid\_u\_maze, ant\_u4\_maze, ant\_u5\_maze}.

\subsection{Python Environment Differences}

\begin{figure}[h]
    \centering
    \includegraphics[width=.5\linewidth]{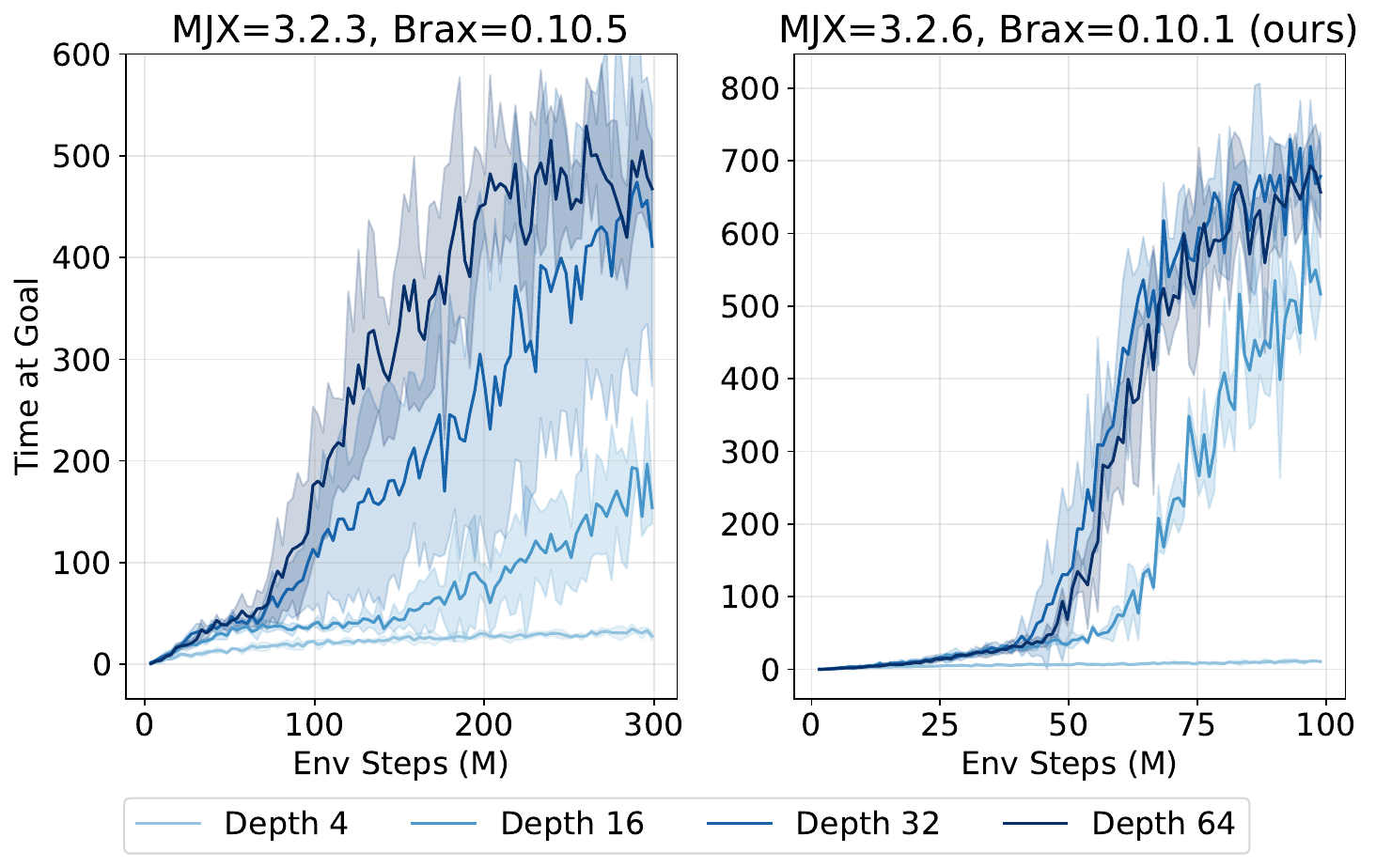}
    \caption{Scaling behavior for humanoid in two different python environments: MJX=3.2.3, Brax=0.10.5 and MJX=3.2.6, Brax=0.10.1 (ours) version of JaxGCRL. Scaling depth improves the performance significantly for both versions. In the environment we used, training requires fewer environment steps to reach a marginally better performance than in other Python environment.}
    \label{fig:physics_diff}
\end{figure}

In all plots presented in the paper, we used MJX 3.2.6 and Brax 0.10.1 to ensure a fair and consistent comparison. During development, we noticed discrepancies in physics behavior between the environment versions we employed (the CleanRL version of JaxGCRL) and the version recommended in a more recent commit of JaxGCRL~\citep{bortkiewicz2024accelerating}. Upon examination, the performance differences (shown in \cref{fig:physics_diff}) stem from a difference in versions in the MJX and Brax packages. Nonetheless, in both sets of MJX and Brax versions, performance scales monotonically with depth.

% In all plots presented in the main paper, we utilized MJX version 3.2.6 and Brax version 0.10.1 to ensure a fair and consistent comparison. During development, we noticed discrepancies in physics behavior between the environment versions we employed (the CleanRL version of JaxGCRL) and the version recommended in a more recent commit of JaxGCRL~\citep{bortkiewicz2024accelerating}. Upon examination, we identified that this variation in learning performance (illustrated in \cref{fig:physics_diff}) is directly attributed to differences between the versions of MJX and Brax. Both packages are actively developed and responsible for defining physics simulations optimized for GPU devices. Importantly, we observed a consistent trend of monotonically increasing performance with depth scaling across both sets of environment versions.

% In all plots in the main paper, we used MJX version 3.2.6 and Brax version 0.10.1, ensuring a fair comparison. During the development, we observed different behavior of physics between the environment versions that we used, \href{https://github.com/MichalBortkiewicz/JaxGCRL/commit/ede9190b4573c091f84c74c265c63e6ed8b73ef8}{CleanRL version of JaxGCRL}, and one recommended for installation in more recent commit of JaxGCRL~\citep{bortkiewicz2024accelerating}. Upon closer examination, this difference in learning performance (\cref{fig:physics_diff}) arises from using different versions of MJX and Brax—two actively developed packages that define physics simulation for GPU devices. Importantly, we observe a similar pattern in the monotonic increase in performance with depth scaling in both versions.

% \section{Wall-clock Time, FLOPs, ...}

% \label{sec:Wall-clock_exps}

\subsection{Wall-clock Time of Our Approach}

We report the wall-clock time of our approach in \cref{tab:Wall-clock_time_transposed}. The table shows results for depths of 4, 8, 16, 32, and 64 across all ten environments, and for the Humanoid U-Maze environment, scaling up to 1024 layers. Overall, wall-clock time increases approximately linearly with depth beyond a certain point.

\vspace{0.25em}

\begin{table}[htbp]
\centering
\caption{Wall-clock time (in hours) for Depth 4, 8, 16, 32, and 64 across all 10 environments.}
\label{tab:Wall-clock_time_transposed}
\resizebox{\textwidth}{!}{
\begin{tabular}{lccccc}
\toprule
\textbf{Environment} & \textbf{Depth 4} & \textbf{Depth 8} & \textbf{Depth 16} & \textbf{Depth 32} & \textbf{Depth 64} \\
\midrule
Humanoid & 1.48 $\pm$ 0.00 & 2.13 $\pm$ 0.01 & 3.40 $\pm$ 0.01 & 5.92 $\pm$ 0.01 & 10.99 $\pm$ 0.01 \\
Ant Big Maze & 2.12 $\pm$ 0.00 & 2.77 $\pm$ 0.00 & 4.04 $\pm$ 0.01 & 6.57 $\pm$ 0.02 & 11.66 $\pm$ 0.03 \\
Ant U4-Maze & 1.98 $\pm$ 0.27 & 2.54 $\pm$ 0.01 & 3.81 $\pm$ 0.01 & 6.35 $\pm$ 0.01 & 11.43 $\pm$ 0.03 \\
Ant U5-Maze & 9.46 $\pm$ 1.75 & 10.99 $\pm$ 0.02 & 16.09 $\pm$ 0.01 & 31.49 $\pm$ 0.34 & 46.40 $\pm$ 0.12 \\
Ant Hardest Maze & 5.11 $\pm$ 0.00 & 6.39 $\pm$ 0.00 & 8.94 $\pm$ 0.01 & 13.97 $\pm$ 0.01 & 23.96 $\pm$ 0.06 \\
Arm Push Easy & 9.97 $\pm$ 1.03 & 11.02 $\pm$ 1.29 & 12.20 $\pm$ 1.43 & 14.94 $\pm$ 1.96 & 19.52 $\pm$ 1.97 \\
Arm Push Hard & 9.74 $\pm$ 1.05 & 10.55 $\pm$ 1.20 & 11.98 $\pm$ 1.49 & 14.40 $\pm$ 1.64 & 18.53 $\pm$ 0.06 \\
Arm Binpick Hard & 18.41 $\pm$ 2.16 & 17.48 $\pm$ 1.88 & 19.47 $\pm$ 0.05 & 21.91 $\pm$ 1.93 & 29.64 $\pm$ 6.10 \\
Humanoid U-Maze & 8.72 $\pm$ 0.01 & 11.29 $\pm$ 0.01 & 16.36 $\pm$ 0.03 & 26.48 $\pm$ 0.05 & 46.74 $\pm$ 0.04 \\
Humanoid Big Maze & 12.45 $\pm$ 0.02 & 15.02 $\pm$ 0.01 & 20.34 $\pm$ 0.01 & 30.61 $\pm$ 0.05 & 50.33 $\pm$ 0.05 \\
\bottomrule
\end{tabular}
}

\end{table}

\begin{table}[htbp]
\centering
\caption{Total wall-clock time (in hours) for training from Depth 4 up to Depth 1024 in the Humanoid U-Maze environment.}
\label{tab:depth_1024_transposed}
\begin{tabular}{lc}
\toprule
\textbf{Depth} & \textbf{Time (h)} \\
\midrule
4 & 3.23 $\pm$ 0.001 \\
8 & 4.19 $\pm$ 0.003 \\
16 & 6.07 $\pm$ 0.003 \\
32 & 9.83 $\pm$ 0.006 \\
64 & 17.33 $\pm$ 0.003 \\
128 & 32.67 $\pm$ 0.124 \\
256 & 73.83 $\pm$ 2.364 \\
512 & 120.88 $\pm$ 2.177 \\
1024 & 134.15 $\pm$ 0.081 \\
\bottomrule
\end{tabular}

\end{table}

\vspace{2em}

\subsection{Wall-clock Time: Comparison to Baselines}

Since the baselines use standard sized networks, naturally our scaled approach incurs higher raw wall-clock time per environment step (\cref{tab:time_comparison_transposed}). However, a more practical metric is the time required to reach a given performance level. As shown in \cref{tab:crl_vs_sac_hours}, our approach outperforms the strongest baseline, SAC, in 7 of 10 environments while requiring less wall-clock time.

\begin{table}[htbp]
\centering
\caption{Wall-clock training time comparison of our method vs. baselines across all 10 environments.}
\label{tab:time_comparison_transposed}
\resizebox{\textwidth}{!}{
\begin{tabular}{lcccccc}
\toprule
\textbf{Environment} & \textbf{Scaled CRL } & \textbf{SAC} & \textbf{SAC+HER} & \textbf{TD3} & \textbf{GCSL} & \textbf{GCBC} \\
\midrule
Humanoid & 11.0 $\pm$ 0.0 & 0.5 $\pm$ 0.0 & 0.6 $\pm$ 0.0 & 0.8 $\pm$ 0.0 & 0.4 $\pm$ 0.0 & 0.6 $\pm$ 0.0 \\
Ant Big Maze & 11.7 $\pm$ 0.0 & 1.6 $\pm$ 0.0 & 1.6 $\pm$ 0.0 & 1.7 $\pm$ 0.0 & 1.5 $\pm$ 0.3 & 1.4 $\pm$ 0.1 \\
Ant U4-Maze & 11.4 $\pm$ 0.0 & 1.2 $\pm$ 0.0 & 1.3 $\pm$ 0.0 & 1.3 $\pm$ 0.0 & 0.7 $\pm$ 0.0 & 1.1 $\pm$ 0.1 \\
Ant U5-Maze & 46.4 $\pm$ 0.1 & 5.7 $\pm$ 0.0 & 6.1 $\pm$ 0.0 & 6.2 $\pm$ 0.0 & 2.8 $\pm$ 0.1 & 5.6 $\pm$ 0.5 \\
Ant Hardest Maze & 24.0 $\pm$ 0.0 & 4.3 $\pm$ 0.0 & 4.5 $\pm$ 0.0 & 5.0 $\pm$ 0.0 & 2.1 $\pm$ 0.6 & 4.4 $\pm$ 0.5 \\
Arm Push Easy & 19.5 $\pm$ 0.6 & 8.3 $\pm$ 0.0 & 8.5 $\pm$ 0.0 & 8.4 $\pm$ 0.0 & 6.4 $\pm$ 0.1 & 8.3 $\pm$ 0.3 \\
Arm Push Hard & 18.5 $\pm$ 0.0 & 8.5 $\pm$ 0.0 & 8.6 $\pm$ 0.0 & 8.3 $\pm$ 0.1 & 5.2 $\pm$ 0.3 & 7.4 $\pm$ 0.5 \\
Arm Binpick Hard & 29.6 $\pm$ 1.3 & 20.7 $\pm$ 0.1 & 20.7 $\pm$ 0.0 & 18.4 $\pm$ 0.3 & 8.0 $\pm$ 0.9 & 16.2 $\pm$ 0.4 \\
Humanoid U-Maze & 46.7 $\pm$ 0.0 & 3.0 $\pm$ 0.0 & 3.5 $\pm$ 0.0 & 5.4 $\pm$ 0.0 & 3.1 $\pm$ 0.1 & 7.2 $\pm$ 0.8 \\
Humanoid Big Maze & 50.3 $\pm$ 0.0 & 8.6 $\pm$ 0.0 & 9.3 $\pm$ 0.0 & 7.5 $\pm$ 1.1 & 5.1 $\pm$ 0.0 & 11.4 $\pm$ 1.9 \\
\bottomrule
\end{tabular}
}
\end{table}

\begin{table}[htbp]
\centering
\caption{Wall-clock time (in hours) for our approach to surpass SAC's final performance. As shown, our approach surpasses SAC performance in less wall-clock time in 7 out of 10 environments. The N/A* entries are because in those environments, scaled CRL doesn’t outperform SAC.}
\label{tab:crl_vs_sac_hours}
\begin{tabular}{lcc}
\toprule
\textbf{Environment} & \textbf{SAC} & \textbf{Scaled CRL (Depth 64)} \\
\midrule
Humanoid & \textbf{0.46} & 6.37 \\
Ant Big Maze & 1.55 & \textbf{0.00} \\
Ant U4-Maze & 1.16 & \textbf{0.00} \\
Ant U5-Maze & 5.73 & \textbf{0.00} \\
Ant Hardest Maze & 4.33 & \textbf{0.45} \\
Arm Push Easy & 8.32 & \textbf{1.91} \\
Arm Push Hard & 8.50 & \textbf{6.65} \\
Arm Binpick Hard & 20.70 & \textbf{4.43} \\
Humanoid U-Maze & \textbf{3.04} & N/A* \\
Humanoid Big Maze & \textbf{8.55} & N/A* \\
\bottomrule
\end{tabular}

\end{table}

\begin{table}[h]
	\centering
	\caption{Hyperparameters}
	\begin{tabular}{cc}
		\toprule
		\textbf{Hyperparameter}                            & \textbf{Value}                 \\
		\midrule
		\verb|num_timesteps|                               & 100M-400M (varying across tasks)                    \\
		\verb|update-to-data (UTD) ratio |                             & 1:40                         \\
        \verb|max_replay_size|                             & 10,000                         \\
		\verb|min_replay_size|                             & 1,000                          \\
		\texttt{episode\_length}
		                                                   & 1,000          \\
		\verb|discounting|                                 & 0.99                           \\
		\verb|num_envs|                                    & 512 \\
		\verb|batch_size|                                  & 512                                                   \\
		% \verb|action_repeat|                               & 1                              \\
		\verb|policy_lr|                                   & 3e-4                           \\
		\verb|critic_lr|                                   & 3e-4                           \\
		% \verb|activation function|                                   & Swish                           \\
		\verb|contrastive_loss_function|                   & \verb|InfoNCE|       \\
		\verb|energy_function|                             & \verb|L2|                      \\
		\verb|logsumexp_penalty|                           & 0.1                            \\
		\verb|Network depth| & depends on the experiment   
        \\
		\verb|Network width| & depends on the experiment   \\
		\verb|representation dimension|                    & 64                             \\
		\bottomrule
	\end{tabular}
	\label{table:params}
\end{table}

% \begin{table}[htbp]
% \centering
% \resizebox{\textwidth}{!}{
% \begin{tabular}{lccccccccc}
% \toprule
% \textbf{Depth} & \textbf{4} & \textbf{8} & \textbf{16} & \textbf{32} & \textbf{64} & \textbf{128} & \textbf{256} & \textbf{512} & \textbf{1024} \\
% \midrule
% \textbf{Time (h)} & 3.23 $\pm$ 0.001 & 4.19 $\pm$ 0.003 & 6.07 $\pm$ 0.003 & 9.83 $\pm$ 0.006 & 17.33 $\pm$ 0.003 & 32.67 $\pm$ 0.1 & 73.83 $\pm$ 2.4 & 120.88 $\pm$ 2.2 & 134.15 $\pm$ 0.08 \\
% \bottomrule
% \end{tabular}
% }
% \caption{Wall-clock time (in hours) for training from Depth 4 up to Depth 1024 in the Humanoid U-Maze environment.}
% \label{tab:depth_1024}
% \end{table}

% \begin{tabular}{c||ccc||ccc||ccc}
% \toprule
%  & \multicolumn{3}{c||}{\textbf{Steps to reach ≥200 success}} 
%  & \multicolumn{3}{c||}{\textbf{Steps to reach ≥400 success}} 
%  & \multicolumn{3}{c}{\textbf{Steps to reach ≥600 success}} \\
% \cmidrule{2-10}
% \textbf{Depth} & 4 & 16 & 32 & 4 & 16 & 32 & 4 & 16 & 32 \\
% \midrule
% \textbf{With} & -- & \textbf{50} & \textbf{42} & -- & \textbf{62} & \textbf{48} & -- & \textbf{77} & \textbf{67} \\
% \textbf{Without} & -- & 64 & 54 & -- & 75 & 64 & -- & -- & 77 \\
% \bottomrule
% \end{tabular}

\newpage
\section*{NeurIPS Paper Checklist}

\begin{enumerate}

\item {\bf Claims}
    \item[] Question: Do the main claims made in the abstract and introduction accurately reflect the paper's contributions and scope?
    \item[] Answer: \answerYes{} % Replace by \answerYes{}, \answerNo{}, or \answerNA{}.
    \item[] Justification: The abstract contains 3 main claims: (1) Depth scaled to 1024 layers; (2) Performance increases 2-50x on CRL and outperforms other goal-conditioned baselines. (3) These performance gains leads to qualitatively new learned behaviors. Each of these claims are clearly substantiated in the main text in Section 4.
    \item[] Guidelines:
    \begin{itemize}
        \item The answer NA means that the abstract and introduction do not include the claims made in the paper.
        \item The abstract and/or introduction should clearly state the claims made, including the contributions made in the paper and important assumptions and limitations. A No or NA answer to this question will not be perceived well by the reviewers. 
        \item The claims made should match theoretical and experimental results, and reflect how much the results can be expected to generalize to other settings. 
        \item It is fine to include aspirational goals as motivation as long as it is clear that these goals are not attained by the paper. 
    \end{itemize}

\item {\bf Limitations}
    \item[] Question: Does the paper discuss the limitations of the work performed by the authors?
    \item[] Answer: \answerYes{} % Replace by \answerYes{}, \answerNo{}, or \answerNA{}.
    \item[] Justification: We included a Limitations section that describes the main limitation of our paper, which is latency of deep networks. We also multiple times in the paper demarcated where our research can be extended by future work.
    \item[] Guidelines:
    \begin{itemize}
        \item The answer NA means that the paper has no limitation while the answer No means that the paper has limitations, but those are not discussed in the paper. 
        \item The authors are encouraged to create a separate "Limitations" section in their paper.
        \item The paper should point out any strong assumptions and how robust the results are to violations of these assumptions (e.g., independence assumptions, noiseless settings, model well-specification, asymptotic approximations only holding locally). The authors should reflect on how these assumptions might be violated in practice and what the implications would be.
        \item The authors should reflect on the scope of the claims made, e.g., if the approach was only tested on a few datasets or with a few runs. In general, empirical results often depend on implicit assumptions, which should be articulated.
        \item The authors should reflect on the factors that influence the performance of the approach. For example, a facial recognition algorithm may perform poorly when image resolution is low or images are taken in low lighting. Or a speech-to-text system might not be used reliably to provide closed captions for online lectures because it fails to handle technical jargon.
        \item The authors should discuss the computational efficiency of the proposed algorithms and how they scale with dataset size.
        \item If applicable, the authors should discuss possible limitations of their approach to address problems of privacy and fairness.
        \item While the authors might fear that complete honesty about limitations might be used by reviewers as grounds for rejection, a worse outcome might be that reviewers discover limitations that aren't acknowledged in the paper. The authors should use their best judgment and recognize that individual actions in favor of transparency play an important role in developing norms that preserve the integrity of the community. Reviewers will be specifically instructed to not penalize honesty concerning limitations.
    \end{itemize}

\item {\bf Theory assumptions and proofs}
    \item[] Question: For each theoretical result, does the paper provide the full set of assumptions and a complete (and correct) proof?
    \item[] Answer: \answerNA{} % Replace by \answerYes{}, \answerNo{}, or \answerNA{}.
    \item[] Justification: This is an empirical paper. As such, no theoretical results that require assumptions or proofs.
    \item[] Guidelines:
    \begin{itemize}
        \item The answer NA means that the paper does not include theoretical results. 
        \item All the theorems, formulas, and proofs in the paper should be numbered and cross-referenced.
        \item All assumptions should be clearly stated or referenced in the statement of any theorems.
        \item The proofs can either appear in the main paper or the supplemental material, but if they appear in the supplemental material, the authors are encouraged to provide a short proof sketch to provide intuition. 
        \item Inversely, any informal proof provided in the core of the paper should be complemented by formal proofs provided in appendix or supplemental material.
        \item Theorems and Lemmas that the proof relies upon should be properly referenced. 
    \end{itemize}

    \item {\bf Experimental result reproducibility}
    \item[] Question: Does the paper fully disclose all the information needed to reproduce the main experimental results of the paper to the extent that it affects the main claims and/or conclusions of the paper (regardless of whether the code and data are provided or not)?
    \item[] Answer: \answerYes{} % Replace by \answerYes{}, \answerNo{}, or \answerNA{}.
    \item[] Justification: Yes, documentation for reproducing the experiments is included alongside the anonymous code.
    \item[] Guidelines:
    \begin{itemize}
        \item The answer NA means that the paper does not include experiments.
        \item If the paper includes experiments, a No answer to this question will not be perceived well by the reviewers: Making the paper reproducible is important, regardless of whether the code and data are provided or not.
        \item If the contribution is a dataset and/or model, the authors should describe the steps taken to make their results reproducible or verifiable. 
        \item Depending on the contribution, reproducibility can be accomplished in various ways. For example, if the contribution is a novel architecture, describing the architecture fully might suffice, or if the contribution is a specific model and empirical evaluation, it may be necessary to either make it possible for others to replicate the model with the same dataset, or provide access to the model. In general. releasing code and data is often one good way to accomplish this, but reproducibility can also be provided via detailed instructions for how to replicate the results, access to a hosted model (e.g., in the case of a large language model), releasing of a model checkpoint, or other means that are appropriate to the research performed.
        \item While NeurIPS does not require releasing code, the conference does require all submissions to provide some reasonable avenue for reproducibility, which may depend on the nature of the contribution. For example
        \begin{enumerate}
            \item If the contribution is primarily a new algorithm, the paper should make it clear how to reproduce that algorithm.
            \item If the contribution is primarily a new model architecture, the paper should describe the architecture clearly and fully.
            \item If the contribution is a new model (e.g., a large language model), then there should either be a way to access this model for reproducing the results or a way to reproduce the model (e.g., with an open-source dataset or instructions for how to construct the dataset).
            \item We recognize that reproducibility may be tricky in some cases, in which case authors are welcome to describe the particular way they provide for reproducibility. In the case of closed-source models, it may be that access to the model is limited in some way (e.g., to registered users), but it should be possible for other researchers to have some path to reproducing or verifying the results.
        \end{enumerate}
    \end{itemize}

\item {\bf Open access to data and code}
    \item[] Question: Does the paper provide open access to the data and code, with sufficient instructions to faithfully reproduce the main experimental results, as described in supplemental material?
    \item[] Answer: \answerYes{} % Replace by \answerYes{}, \answerNo{}, or \answerNA{}.
    \item[] Justification: See link to anonymous code in Abstract.
    \item[] Guidelines:
    \begin{itemize}
        \item The answer NA means that paper does not include experiments requiring code.
        \item Please see the NeurIPS code and data submission guidelines (\url{https://nips.cc/public/guides/CodeSubmissionPolicy}) for more details.
        \item While we encourage the release of code and data, we understand that this might not be possible, so “No” is an acceptable answer. Papers cannot be rejected simply for not including code, unless this is central to the contribution (e.g., for a new open-source benchmark).
        \item The instructions should contain the exact command and environment needed to run to reproduce the results. See the NeurIPS code and data submission guidelines (\url{https://nips.cc/public/guides/CodeSubmissionPolicy}) for more details.
        \item The authors should provide instructions on data access and preparation, including how to access the raw data, preprocessed data, intermediate data, and generated data, etc.
        \item The authors should provide scripts to reproduce all experimental results for the new proposed method and baselines. If only a subset of experiments are reproducible, they should state which ones are omitted from the script and why.
        \item At submission time, to preserve anonymity, the authors should release anonymized versions (if applicable).
        \item Providing as much information as possible in supplemental material (appended to the paper) is recommended, but including URLs to data and code is permitted.
    \end{itemize}

\item {\bf Experimental setting/details}
    \item[] Question: Does the paper specify all the training and test details (e.g., data splits, hyperparameters, how they were chosen, type of optimizer, etc.) necessary to understand the results?
    \item[] Answer: \answerYes{} % Replace by \answerYes{}, \answerNo{}, or \answerNA{}.
    \item[] Justification: See Experiments section and Appendix on Experimental Details
    \item[] Guidelines:
    \begin{itemize}
        \item The answer NA means that the paper does not include experiments.
        \item The experimental setting should be presented in the core of the paper to a level of detail that is necessary to appreciate the results and make sense of them.
        \item The full details can be provided either with the code, in appendix, or as supplemental material.
    \end{itemize}

\item {\bf Experiment statistical significance}
    \item[] Question: Does the paper report error bars suitably and correctly defined or other appropriate information about the statistical significance of the experiments?
    \item[] Answer: \answerYes{} % Replace by \answerYes{}, \answerNo{}, or \answerNA{}.
    \item[] Justification: Error bars in figures depict one standard error across random seeds. We used 5 seeds in Figure 1. For other figures in the main text, we could only run 3 seeds because of computational constraints.
    \item[] Guidelines:
    \begin{itemize}
        \item The answer NA means that the paper does not include experiments.
        \item The authors should answer "Yes" if the results are accompanied by error bars, confidence intervals, or statistical significance tests, at least for the experiments that support the main claims of the paper.
        \item The factors of variability that the error bars are capturing should be clearly stated (for example, train/test split, initialization, random drawing of some parameter, or overall run with given experimental conditions).
        \item The method for calculating the error bars should be explained (closed form formula, call to a library function, bootstrap, etc.)
        \item The assumptions made should be given (e.g., Normally distributed errors).
        \item It should be clear whether the error bar is the standard deviation or the standard error of the mean.
        \item It is OK to report 1-sigma error bars, but one should state it. The authors should preferably report a 2-sigma error bar than state that they have a 96\% CI, if the hypothesis of Normality of errors is not verified.
        \item For asymmetric distributions, the authors should be careful not to show in tables or figures symmetric error bars that would yield results that are out of range (e.g. negative error rates).
        \item If error bars are reported in tables or plots, The authors should explain in the text how they were calculated and reference the corresponding figures or tables in the text.
    \end{itemize}

\item {\bf Experiments compute resources}
    \item[] Question: For each experiment, does the paper provide sufficient information on the computer resources (type of compute workers, memory, time of execution) needed to reproduce the experiments?
    \item[] Answer: \answerYes{} % Replace by \answerYes{}, \answerNo{}, or \answerNA{}.
    \item[] Justification: Compute resources are detailed in the appendix.
    \item[] Guidelines:
    \begin{itemize}
        \item The answer NA means that the paper does not include experiments.
        \item The paper should indicate the type of compute workers CPU or GPU, internal cluster, or cloud provider, including relevant memory and storage.
        \item The paper should provide the amount of compute required for each of the individual experimental runs as well as estimate the total compute. 
        \item The paper should disclose whether the full research project required more compute than the experiments reported in the paper (e.g., preliminary or failed experiments that didn't make it into the paper). 
    \end{itemize}
    
\item {\bf Code of ethics}
    \item[] Question: Does the research conducted in the paper conform, in every respect, with the NeurIPS Code of Ethics \url{https://neurips.cc/public/EthicsGuidelines}?
    \item[] Answer: \answerYes{} % Replace by \answerYes{}, \answerNo{}, or \answerNA{}.
    \item[] Justification: No known violations of the Code of Ethics.
    \item[] Guidelines:
    \begin{itemize}
        \item The answer NA means that the authors have not reviewed the NeurIPS Code of Ethics.
        \item If the authors answer No, they should explain the special circumstances that require a deviation from the Code of Ethics.
        \item The authors should make sure to preserve anonymity (e.g., if there is a special consideration due to laws or regulations in their jurisdiction).
    \end{itemize}

\item {\bf Broader impacts}
    \item[] Question: Does the paper discuss both potential positive societal impacts and negative societal impacts of the work performed?
    \item[] Answer: \answerYes{} % Replace by \answerYes{}, \answerNo{}, or \answerNA{}.
    \item[] Justification: The Conclusion notes that there are no immediately societal impacts of the work.
    \item[] Guidelines:
    \begin{itemize}
        \item The answer NA means that there is no societal impact of the work performed.
        \item If the authors answer NA or No, they should explain why their work has no societal impact or why the paper does not address societal impact.
        \item Examples of negative societal impacts include potential malicious or unintended uses (e.g., disinformation, generating fake profiles, surveillance), fairness considerations (e.g., deployment of technologies that could make decisions that unfairly impact specific groups), privacy considerations, and security considerations.
        \item The conference expects that many papers will be foundational research and not tied to particular applications, let alone deployments. However, if there is a direct path to any negative applications, the authors should point it out. For example, it is legitimate to point out that an improvement in the quality of generative models could be used to generate deepfakes for disinformation. On the other hand, it is not needed to point out that a generic algorithm for optimizing neural networks could enable people to train models that generate Deepfakes faster.
        \item The authors should consider possible harms that could arise when the technology is being used as intended and functioning correctly, harms that could arise when the technology is being used as intended but gives incorrect results, and harms following from (intentional or unintentional) misuse of the technology.
        \item If there are negative societal impacts, the authors could also discuss possible mitigation strategies (e.g., gated release of models, providing defenses in addition to attacks, mechanisms for monitoring misuse, mechanisms to monitor how a system learns from feedback over time, improving the efficiency and accessibility of ML).
    \end{itemize}
    
\item {\bf Safeguards}
    \item[] Question: Does the paper describe safeguards that have been put in place for responsible release of data or models that have a high risk for misuse (e.g., pretrained language models, image generators, or scraped datasets)?
    \item[] Answer: \answerNA{} % Replace by \answerYes{}, \answerNo{}, or \answerNA{}.
    \item[] Justification: No immediate impact to high-risk applications. 
    \item[] Guidelines:
    \begin{itemize}
        \item The answer NA means that the paper poses no such risks.
        \item Released models that have a high risk for misuse or dual-use should be released with necessary safeguards to allow for controlled use of the model, for example by requiring that users adhere to usage guidelines or restrictions to access the model or implementing safety filters. 
        \item Datasets that have been scraped from the Internet could pose safety risks. The authors should describe how they avoided releasing unsafe images.
        \item We recognize that providing effective safeguards is challenging, and many papers do not require this, but we encourage authors to take this into account and make a best faith effort.
    \end{itemize}

\item {\bf Licenses for existing assets}
    \item[] Question: Are the creators or original owners of assets (e.g., code, data, models), used in the paper, properly credited and are the license and terms of use explicitly mentioned and properly respected?
    \item[] Answer: \answerNA{} % Replace by \answerYes{}, \answerNo{}, or \answerNA{}.
    \item[] Justification: Benchmarks used are appropriately cited in the main text.
    \item[] Guidelines:
    \begin{itemize}
        \item The answer NA means that the paper does not use existing assets.
        \item The authors should cite the original paper that produced the code package or dataset.
        \item The authors should state which version of the asset is used and, if possible, include a URL.
        \item The name of the license (e.g., CC-BY 4.0) should be included for each asset.
        \item For scraped data from a particular source (e.g., website), the copyright and terms of service of that source should be provided.
        \item If assets are released, the license, copyright information, and terms of use in the package should be provided. For popular datasets, \url{paperswithcode.com/datasets} has curated licenses for some datasets. Their licensing guide can help determine the license of a dataset.
        \item For existing datasets that are re-packaged, both the original license and the license of the derived asset (if it has changed) should be provided.
        \item If this information is not available online, the authors are encouraged to reach out to the asset's creators.
    \end{itemize}

\item {\bf New assets}
    \item[] Question: Are new assets introduced in the paper well documented and is the documentation provided alongside the assets?
    \item[] Answer: \answerNA{} % Replace by \answerYes{}, \answerNo{}, or \answerNA{}.
    \item[] Justification: Datasets and benchmark used are all from prior work and appropriately cited.
    \item[] Guidelines:
    \begin{itemize}
        \item The answer NA means that the paper does not release new assets.
        \item Researchers should communicate the details of the dataset/code/model as part of their submissions via structured templates. This includes details about training, license, limitations, etc. 
        \item The paper should discuss whether and how consent was obtained from people whose asset is used.
        \item At submission time, remember to anonymize your assets (if applicable). You can either create an anonymized URL or include an anonymized zip file.
    \end{itemize}

\item {\bf Crowdsourcing and research with human subjects}
    \item[] Question: For crowdsourcing experiments and research with human subjects, does the paper include the full text of instructions given to participants and screenshots, if applicable, as well as details about compensation (if any)? 
    \item[] Answer: \answerNA{} % Replace by \answerYes{}, \answerNo{}, or \answerNA{}.
    \item[] Justification: No crowdsourcing experiments.
    \item[] Guidelines:
    \begin{itemize}
        \item The answer NA means that the paper does not involve crowdsourcing nor research with human subjects.
        \item Including this information in the supplemental material is fine, but if the main contribution of the paper involves human subjects, then as much detail as possible should be included in the main paper. 
        \item According to the NeurIPS Code of Ethics, workers involved in data collection, curation, or other labor should be paid at least the minimum wage in the country of the data collector. 
    \end{itemize}

\item {\bf Institutional review board (IRB) approvals or equivalent for research with human subjects}
    \item[] Question: Does the paper describe potential risks incurred by study participants, whether such risks were disclosed to the subjects, and whether Institutional Review Board (IRB) approvals (or an equivalent approval/review based on the requirements of your country or institution) were obtained?
    \item[] Answer: \answerNA{} % Replace by \answerYes{}, \answerNo{}, or \answerNA{}.
    \item[] Justification: No human subject experiments
    \item[] Guidelines:
    \begin{itemize}
        \item The answer NA means that the paper does not involve crowdsourcing nor research with human subjects.
        \item Depending on the country in which research is conducted, IRB approval (or equivalent) may be required for any human subjects research. If you obtained IRB approval, you should clearly state this in the paper. 
        \item We recognize that the procedures for this may vary significantly between institutions and locations, and we expect authors to adhere to the NeurIPS Code of Ethics and the guidelines for their institution. 
        \item For initial submissions, do not include any information that would break anonymity (if applicable), such as the institution conducting the review.
    \end{itemize}

\item {\bf Declaration of LLM usage}
    \item[] Question: Does the paper describe the usage of LLMs if it is an important, original, or non-standard component of the core methods in this research? Note that if the LLM is used only for writing, editing, or formatting purposes and does not impact the core methodology, scientific rigorousness, or originality of the research, declaration is not required.
    %this research? 
    \item[] Answer: \answerNA{} % Replace by \answerYes{}, \answerNo{}, or \answerNA{}.
    \item[] Justification: LLMs were not used in writing the paper, and were only used for occasional code debugging.
    \item[] Guidelines:
    \begin{itemize}
        \item The answer NA means that the core method development in this research does not involve LLMs as any important, original, or non-standard components.
        \item Please refer to our LLM policy (\url{https://neurips.cc/Conferences/2025/LLM}) for what should or should not be described.
    \end{itemize}

\end{enumerate}

\clearpage

%%%%%%%%%%%%%%%%%%%%%%%%%%%%%%%%%%%%%%%%%%%%%%%%%%%%%%%%%%%%%%%%%%%%%%%%%%%%%%%
%%%%%%%%%%%%%%%%%%%%%%%%%%%%%%%%%%%%%%%%%%%%%%%%%%%%%%%%%%%%%%%%%%%%%%%%%%%%%%%

\end{document}

%% file: math_commands.tex
\usepackage{amsmath,amsfonts,bm,cancel}

\def\eqref#1{equation~\ref{#1}}

\def\1{\bm{1}}

\DeclareMathAlphabet{\mathsfit}{\encodingdefault}{\sfdefault}{m}{sl}
\SetMathAlphabet{\mathsfit}{bold}{\encodingdefault}{\sfdefault}{bx}{n}

\definecolor{positive}{rgb}{0,0.55,0}
\definecolor{negative}{rgb}{0.75,0,0}